\documentclass{article}

\usepackage{PRIMEarxiv}

\usepackage[utf8]{inputenc} % allow utf-8 input
\usepackage[T1]{fontenc}    % use 8-bit T1 fonts
\usepackage{url}            % simple URL typesetting
\usepackage{booktabs}       % professional-quality tables
\usepackage{amsfonts}       % blackboard math symbols
\usepackage{nicefrac}       % compact symbols for 1/2, etc.
\usepackage{microtype}      % microtypography
\usepackage{lipsum}
\usepackage{fancyhdr}       % header
\usepackage{graphicx}       % graphics
\graphicspath{{media/}}     % organize your images and other figures under media/ folder

\usepackage{enumitem}
\usepackage{amsmath}
\usepackage{amsthm}
\usepackage{amssymb}
\usepackage{algorithm}
\usepackage{algorithmic}
\usepackage{multirow}
\usepackage{multicol}
\usepackage{array}
\usepackage{subfigure}

\newcommand{\sysname}{MLDGG}

\newtheorem{lemma}{Lemma}
\newtheorem{theorem}{Theorem}

\title{MLDGG: Meta-Learning for Domain Generalization on Graphs
}

\author{
  Qin Tian$^1$, Chen Zhao$^2$, Minglai Shao$^3$, Wenjun Wang$^1$, Yujie Lin$^3$, Dong Li$^2$ \\
  $^1$College of Intelligence and Computing, Tianjin University \\
  $^2$Department of Computer Science, Baylor University\\
  $^3$School of New Media and Communication, Tianjin University \\
  \texttt{\{tianqin123,shaoml,linyujie\_22\}@tju.edu.cn, \{chen\_zhao,dong\_li1\}@baylor.edu} \\
}

\begin{document}
\maketitle

\begin{abstract}
Domain generalization on graphs aims to develop models with robust generalization capabilities, ensuring effective performance on the testing set despite disparities between testing and training distributions. 
However, existing methods often rely on static encoders directly applied to the target domain, constraining its flexible adaptability. 
% In response, we propose a novel meta-learning approach for graph domain generalization.
In contrast to conventional methodologies, which concentrate on developing specific generalized models, our framework, \sysname{}, endeavors to achieve adaptable generalization across diverse domains by integrating cross-multi-domain meta-learning with structure learning and semantic identification.
Initially, it introduces a generalized structure learner to mitigate the adverse effects of task-unrelated edges, enhancing the comprehensiveness of representations learned by Graph Neural Networks (GNNs) while capturing shared structural information across domains.
% augment the proficiency of the Graph Neural Networks (GNNs) in acquiring a more effective and resilient graph structure and node embeddings. 
Subsequently, a representation learner is designed to disentangle domain-invariant semantic and domain-specific variation information in node embedding by leveraging causal reasoning for semantic identification, further enhancing generalization. 
In the context of meta-learning, meta-parameters for both learners are optimized to facilitate knowledge transfer and enable effective adaptation to graphs through fine-tuning within the target domains, where target graphs are inaccessible during training.
% We conduct three different distribution shift settings to evaluate the performance of domain generalization.
Our empirical results demonstrate that \sysname{} surpasses baseline methods, showcasing the effectiveness in three different distribution shift settings.
% Code is available at: https://anonymous.4open.science/r/meta\_DG-552F/.
\end{abstract}

% keywords can be removed

\section{Introduction}
\label{introduction}

\begin{figure}[t]
    \centering
    \includegraphics[width=\linewidth]{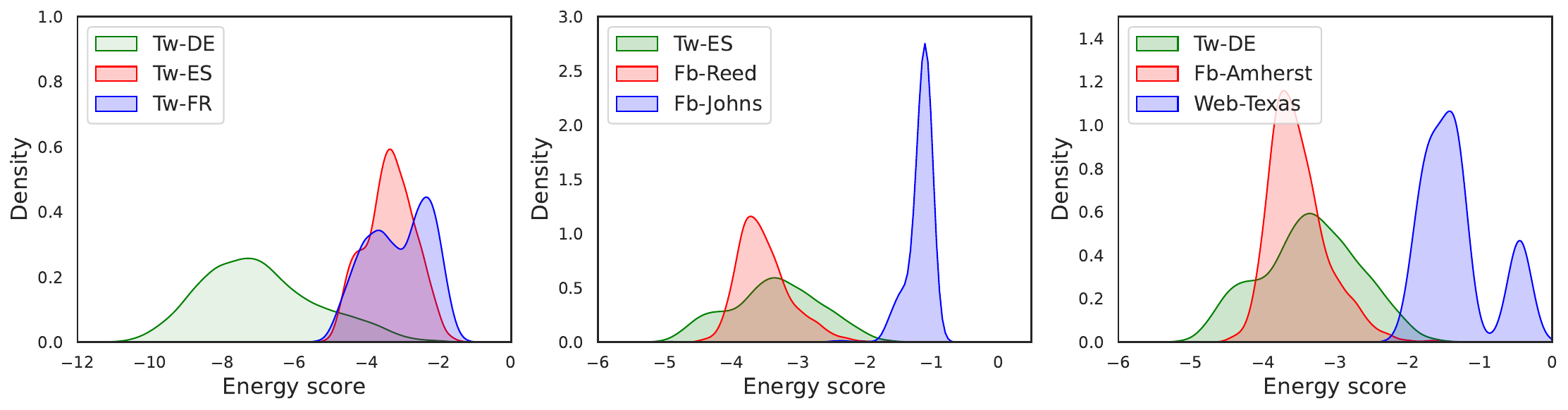}
    % \vspace{-7mm}
    \caption{Visualizations of distribution shifts on graphs demonstrated using energy scores \cite{wu2023gnnsafe} of nodes across different graphs, where each graph is sampled from a distinct domain characterized by variations of node features and topological structures simultaneously. 
    The legend of all sub-figures follows the same naming format, \textit{i.e.,} "DataName - DomainName". 
    (Left) Graphs are sampled from the same dataset \textsc{Twitch}~\cite{rozemberczki2021multi}.
    (Middle and Right) Graphs are sampled from different datasets, \textsc{Twitch}, \textsc{Fb-100}~\cite{traud2012social}, and \textsc{WebKB}~\cite{pei2020geom}.
    }
    \label{fig:energy_score}
    % \vspace{-3mm}
\end{figure}

Domain generalization is a fundamental research area in machine learning that aims to enhance the ability of models learned from source domains to generalize well to different target domains ~\cite{muandet2013domain,li2018domain,shao2024supervised,zhao2024algorithmic,zhao2024dynamic,li2024learning,lin2024towards,lin2023adaptation,zhao2023towards,zhao2022adaptive,zhao2021fairness,jiang2024feed}.
While handling distribution shifts across domains on Euclidean data has achieved significant success~\cite{mahajan2021domain,lv2022causality}, there has been limited focus on graph-structured data due to specific challenges where domains are characterized by variations of node features and graph topological structures simultaneously.
% For example, in a protein-protein interaction network, there are significant changes in the distribution of features and interactive patterns (\textit{i.e.,} topological structures) among proteins from different species (\textit{i.e.,} domain)~\cite{gene2019gene}.
Fig.~\ref{fig:energy_score} illustrates the presence of distribution disparities on graphs, with each graph sampled from a distinct domain. 
Consequently, a model trained on one graph domain (\textit{e.g.,} gamer networks, \textsc{Twitch}) may exhibit poor generalization performance when deployed in a different domain (\textit{e.g.,} social networks, \textsc{FB-100}).

% As shown in Fig.~\ref{fig:energy_score}, there are significant distribution disparities among three networks constructed by different relationships: \textsc{Twitch} (game relationships), \textsc{FB-100} (social relationships), and \textsc{WebKB} (hyperlink relationships).

To address the problem of domain generalization on graphs, several efforts have been made.
Existing approaches on invariant learning with Graph Neural Networks (GNNs)~\cite{liu2023flood,wu2022handling} focus on encoding invariant information of graphs by minimizing the risk across various environments under the assumption that the information determining labels remain constant.
They usually assume access to abundant and diverse training domains, prompting researchers to propose data augmentation~\cite{park2021metropolis,kong2022robust,sui2022adversarial} to alleviate the problem, which strives to diversify the training domains as much as possible to improve the generalization ability of the model.
However, an overly flexible domain augmentation strategy can create implausible augmented domains~\cite{du2020learning}.
Additionally, the complexity of real-world graph structures and the often unknown underlying data generation mechanisms make it challenging to acquire the knowledge necessary for generating new graphs.
% \textcolor{red}{
% [one more sentence goes here....]
% }

Moreover, to alleviate the above obstacles and enhance the interpretability of generalized models, causal reasoning is often combined with invariant learning~\cite{yang2021causalvae,sui2022causal}.
The invariance principle from causality elucidates and models the underlying generative processes of graphs, targeting the identification of stable causal relationships across different domains.
Nevertheless, studies show that trained GNNs are heavily biased towards specific graph structures and cannot effectively address the domain variations on graph topology structures~\cite{li2022learning}.

% This method focuses on encoding invariant information of graphs by minimizing the risk across various environments under the assumption that the information determining labels remain constant. 
% They usually assume access to abundant and diverse training domains, prompting researchers to propose data augmentation techniques~\cite{park2021metropolis,kong2022robust,sui2022adversarial} to alleviate this problem, which strives to diversify the training domains as much as possible to improve the generalization ability of the model.
% Unfortunately, an overly flexible domain augmentation strategy can create implausible augmented domains~\cite{du2020learning}.

% Furthermore, to enhance the interpretability of generalized models, causal reasoning~\cite{yang2021causalvae,sui2022causal} has been proposed and is often combined with invariant learning.
% The invariance principle from causality elucidates and models the underlying generative processes of graphs, targeting the identification of stable causal relationships across different domains.
% The methods discussed integrate node attributes and graph topology with GNNs to capture invariant information across domains.
% Nevertheless, studies show that trained GNNs are heavily biased towards specific graph structures and cannot effectively address the topology structural shift problem~\cite{li2022learning}.

Additionally, some studies integrate structure learning to improve the robustness of generalized GNNs, such as capturing the domain-independent and domain-invariant clusters to learn invariant representations~\cite{li2022learning} by training static encoders shared by all source graphs~\cite{zhao2023graphglow}.
However, capturing invariant topology structure learners across domains reduces the adaptability of the structural encoder and limits its ability to accommodate various distributions.
% In addition, the above methods employ a static encoder trained from multiple source domains to the target domain, limiting their adaptability to the various distributions.
% The distribution of graph data generated by different mechanisms varies greatly in the real world. 
Hence, it is urgent to train models who possess transferable knowledge across domains with various distribution shifts.

In this paper, we propose a novel cross-multi-domain meta-learning framework, \sysname{}, designed to acquire transferable knowledge from graphs sampled in the source domain and generalize to those in the target domain, where target graphs are inaccessible during training.
% We focus on capturing the invariant patterns of the truly predictive properties in different distribution domains.
Specifically, to address the problem of node-level domain generalization on graphs, where domain variations are characterized by graph topological structures and node features, \sysname{} comprises two key components: a structure learner and a representation learner.
% \sysname{} consists of a structure learner and a representation learner within the context of meta-learning to facilitate knowledge transfer between source and target graphs.
The structure learner aims to mitigate the adverse effects of task-unrelated edges and capture structure knowledge shared across different domains, enhancing the comprehensiveness of representations learned by GNNs. 
The representation learner disentangles semantic and variation factors to capture the invariant patterns of the truly predicting properties in different domains.
% It first incorporates a structured learner to bolster the proficiency of GNNs in acquiring more effective and resilient graph structures and node embeddings.
% Then, a representation learner is employed to mitigate the pseudo-correlation between semantic and domain-specific information in node embeddings, thereby enhancing generalization. 
In the context of meta-learning, the goal of \sysname{} aims to learn optimal meta-parameters (initialization) for both learners so that they can facilitate knowledge transfer and enable effective adaptation to graphs through fine-tuning within the target domains.
Our contributions are summarized as follows:
\begin{itemize}[leftmargin=*]
    \item We propose a novel cross-multi-domain meta-learning framework on graphs. It is designed to acquire transferable knowledge from graphs sampled in the source domain and generalize to those in the target domain, where target graphs are inaccessible during training.
    \item The framework consists of two key learners: a structure learner, which captures shared topology patterns across different graph domains to enhance the robustness of GNNs, and a representation learner, which disentangles domain-invariant semantics from domain-specific variations.
    In the context of meta-learning, the parameter initializations for both learners are optimized to facilitate knowledge transfer and enable effective adaptation to graphs through fine-tuning within the target domains.
    % The structure learner facilitates the model by optimizing the GNNs to acquire more intricate and conceptually rich node representations. Consequtly, the representation learner directs the model to disentangle the semantic and domain-specific variation factor within node representations.
    \item Empirically, we conduct three distinct cross-domain settings to assess the generalization ability of \sysname{} for node-level prediction tasks under different degrees of distribution shifts using real-world graph datasets. Our method consistently outperforms state-of-the-art baseline approaches.
\end{itemize}

\section{Related Work}
%\subsection{Domain Generalization on Graphs}
\textbf{Domain Generalization on Graphs.} Domain generalization aims to generalize a model trained on multiple seen domains with diverse distributions to perform well on an unseen target domain~\cite{robey2021model}.
The study of domain generalization on graphs presents significant challenges due to the irregular nature of graph data and the complex dependencies between nodes~\cite{garg2021learn,zhou2022domain}.
Methodologically, various strategies such as robust optimization~\cite{qian2019robust,sagawa2019distributionally}, invariant learning~\cite{krueger2021out,wu2022handling,liu2023flood}, causal  approaches~\cite{yang2021causalvae,sui2022causal} and meta-learning domain generalization~\cite{li2018learning,chen2022discriminative,hassani2022cross,zhang2022adapting} have been employed to tackle this problem.
Robust optimization improves the generalization ability by improving the model's performance in the worst-case data distribution.
Invariant learning minimizes prediction variance across domains to capture invariant features across domains.
Causal approaches are dedicated to separating inclusive information factors (semantic information) and irrelevant factors, utilizing the principles of causal graphs in an unsupervised or semi-supervised manner.
In addition, GraphGlow~\cite{zhao2023graphglow} improves GNN generalization by learning generic graph structures.
It trains a static structure encoder to capture invariant structure information across domains and apply it in the target domain, which reduces the adaptability of the structural encoder
and limits its ability to accommodate various distributions.
% The goal of GraphGlow is to learn a generalized structure learner to implement an optimized representation of GNN.
Despite GNNs' ability to extract abstract representations, they mix the domain-invariant semantic factor with the domain-specific variation factor. 
Meta-learning learns prior experiences during meta-training and transforms learned knowledge to the target domain by simple fine-tuning.
Following similar spirits, we use a combination of learning domain-invariant semantic information and learning-to-learn strategies to achieve domain generalization across domains. 
%\subsection{Meta Learning on Graphs}

\textbf{Meta-Learning on Graphs.} Meta-learning, also known as "learning to learn", focuses on the ability of a model to learn and adapt to new tasks or domains quickly and efficiently~\cite{finn2017model,vilalta2002perspective}. 
It has been widely used in generalization problems~\cite{li2019episodic,balaji2018metareg,dou2019domain,du2020learning}.
Consequently, meta-learning for graphs generally combines the advantages of GNNs and meta-learning to implement generalization on irregular graph data ~\cite{huang2020graph,zhou2019meta}.
From the learning tasks of view, these methods generally fall into three categories, node-level~\cite{yao2020graph,wang2020graph}, edge-level~\cite{huang2020graph,zhang2022adapting} and graph-level~\cite{ma2020adaptive,guo2021few}.
Methodologically, these approaches incorporate metric-based~\cite{sung2018learning,tan2022graph}, which are aimed at learning metrics to quantify the similarity between task-specific support and query sets, and optimization-based methods~\cite{finn2018probabilistic,wang2020generalizing} that concentrate on effectively training a well-initialized learner capable of rapidly adapting to new few-shot tasks via simple fine-tuning.
However, they only consider the scene where all tasks originate from the same domain, the challenge of generalizing prior experiences from cross-multi-domain graphs during meta-training and transferring knowledge to unseen domains is unexplored in the existing literature.

\section{Preliminaries}
We list all notations used in this paper in Table~\ref{tab:detailed_notations} in Appendix~\ref{notations}.

\textbf{Node-level Domain Generalization on Graphs.}
Given a set of graphs $\mathcal{G}=\{ G^{e_i}\}_{i=1}^{|\mathcal{E}|}$, where each graph $G^{e_i} = (A^{e_i}, X^{e_i})$ is sampled from a unique domain $e_i \in \mathcal{E}$ and a domain $e_i$ is defined as a joint distribution $\mathbb{P}(A^{e_i},X^{e_i})$. 
In each graph $G^{e_i}$, we denote ${A}^{e_i} \in \{0,1\}^{|\mathcal{V}^{e_i}| \times |\mathcal{V}^{e_i}|}$ the adjacency matrix, where $\mathcal{V}^{e_i}$ is a collection of nodes. ${X}^{e_i} =\{\mathbf{x}^{e_i}_j\}_{j=1}^{|\mathcal{V}^{e_i}|} \in \mathbb{R}^{|\mathcal{V}^{e_i}| \times D^{e_i}}$ represents the node feature matrix of $D$-dimensional vectors. $\mathbf{y}^{e_i}=\{y_j^{e_i}\}_{j=1}^{|\mathcal{V}^{e_i}|}\in\mathbb{R}^{|\mathcal{V}^{e_i}|}$ denotes node labels in $G^{e_i}$.

For node-level domain generalization, each graph $G^{e_i}\in\mathcal{G}$ is associated with a specific variation in graphic topology $A^{e_i}$ and node features $X^{e_i}$. 
The graph set $\mathcal{G}$ is partitioned into multiple source graphs $\mathcal{G}_s=\{ G^{e_i}\}_{i=1}^{K}$ where $ K = |\mathcal{E}_s|$ and $\mathcal{E}_s\subset\mathcal{E}$, and target graphs $\mathcal{G}_t$ with $\mathcal{E}_t=\mathcal{E}\backslash\mathcal{E}_s$.
The objective is to maintain satisfactory generalization performance in node-level prediction accuracy transitioning from given source graphs $\mathcal{G}_s$ to target graphs $\mathcal{G}_t$, with the condition that $\mathcal{G}_t$ remains inaccessible during training.

\textbf{Meta-Learning}
is an approach aimed at training a model over a range of tasks to be able to adapt to new tasks rapidly \cite{schmidhuber1987evolutionary,qiao2020learning,du2020learning}, 
% to rapidly adapt and generalize across various tasks or domains \textcolor{red}{[citations]}. 
% Unlike traditional learning paradigms focused on mastering specific tasks, meta-learning is devoted to equipping a model with the ability to efficiently acquire new knowledge and adapt its learning process.
where each task $\mathcal{T}^i \sim \mathbb{P}(\mathcal{T})$ associated with a data batch is partitioned into a support set $\mathcal{T}^i_{sup}$ for the learning phase and a query set $\mathcal{T}^i_{qry}$ for evaluation purposes.
% assuming that multiple tasks $\{\mathcal{T}_i\}_{i=1}^{M}$ are sampled from a fixed distribution, where $\mathcal{T}_i \sim \mathbb{P}(\mathcal{T})$.
% The datasets associated with these tasks are accessible. 
% Each task $\mathcal{T}_i$ is partitioned into $\mathcal{T}_i^{sup}$ for the learning phase and $\mathcal{T}_i^{qry}$ for evaluation purposes.
% Typically, $\mathcal{T}_i^{sup}$ has a smaller sample size compared to $\mathcal{T}_i^{qry}$.
Meta-learning can be described as "learning to learn" because it involves finding a meta-parameter $\boldsymbol{\theta}$ from which one can quickly derive multiple optimized parameters $\{\boldsymbol{\theta}'^i\}_{i=1}^M$ specific to individual tasks $\{\mathcal{T}^i\}_{i=1}^M$.

Model-agnostic meta-learning (MAML)~\cite{finn2017model} is a notable gradient-based meta-learning approach that has demonstrated remarkable success in generalization. The core assumption of MAML is that some internal representations are better suited to transfer learning. 
% In the supervised learning setting of MAML, 

During training, the model first learns from $\mathcal{T}^i_{sup}$ for each task $\mathcal{T}^i$ and accordingly optimizes the task-specific parameter to $\boldsymbol{\theta}'^i$ with one or few gradient steps. The meta-parameter $\boldsymbol{\theta}$ is updated through query losses evaluated from $\{\mathcal{T}^i_{qry}\}_{i=1}^M$ based on $\{\boldsymbol{\theta}'^i\}_{i=1}^M$.
\begin{equation}
% \footnotesize
    \boldsymbol{\theta} = \arg\min_{\boldsymbol{\theta}} \frac{1}{M} \sum_{i}^{M} \ell(\boldsymbol{\theta}'^i,\mathcal{T}^i_{qry}),
    \text{ where } \boldsymbol{\theta}'^i =  \boldsymbol{\theta}-\alpha \nabla \ell(\boldsymbol{\theta},\mathcal{T}^i_{sup}),
\end{equation}
where $\ell:\Theta\times\mathbb{R}^d \rightarrow\mathbb{R}$ is the cross-entropy loss for classification and $\alpha>0$ is the learning rate.
The goal of MAML is to learn an effective model initialization $\boldsymbol{\theta}$ using $M$ training tasks, enabling rapid fine-tuning on the support set $\mathcal{T}^t_{sup}$ of the target task $\mathcal{T}^t$ to achieve optimal performance on $\mathcal{T}^{t}_{qry}$, where $\mathcal{T}^t=\{\mathcal{T}^t_{sup},\mathcal{T}^t_{qry}\}$.

Graph generalization using MAML leverages a similar GNN-based task distribution to accumulate transferable knowledge from prior learning experiences. 
However, the original MAML~\cite{finn2017model} assumes that all tasks originate from the same distribution, which hinders its ability to generalize across multiple domains~\cite{antoniou2019train,lin2023multi}.
Additionally, GNNs can introduce noise information from task-unrelated edges, negatively impacting performance. 
As discussed in Section~\ref{introduction}, GNN methods combined with structural optimization typically learn a static structure, which constrains the model's ability to generalize to varying topology distribution shifts.
To overcome these limitations, a cross-multi-domain robust algorithm is required, as we will discuss next.

% It is crucial to acknowledge that MAML operates under the assumption that all tasks stem from the same distribution.

\begin{figure}[t]
    \centering
    \includegraphics[width=0.8\linewidth]{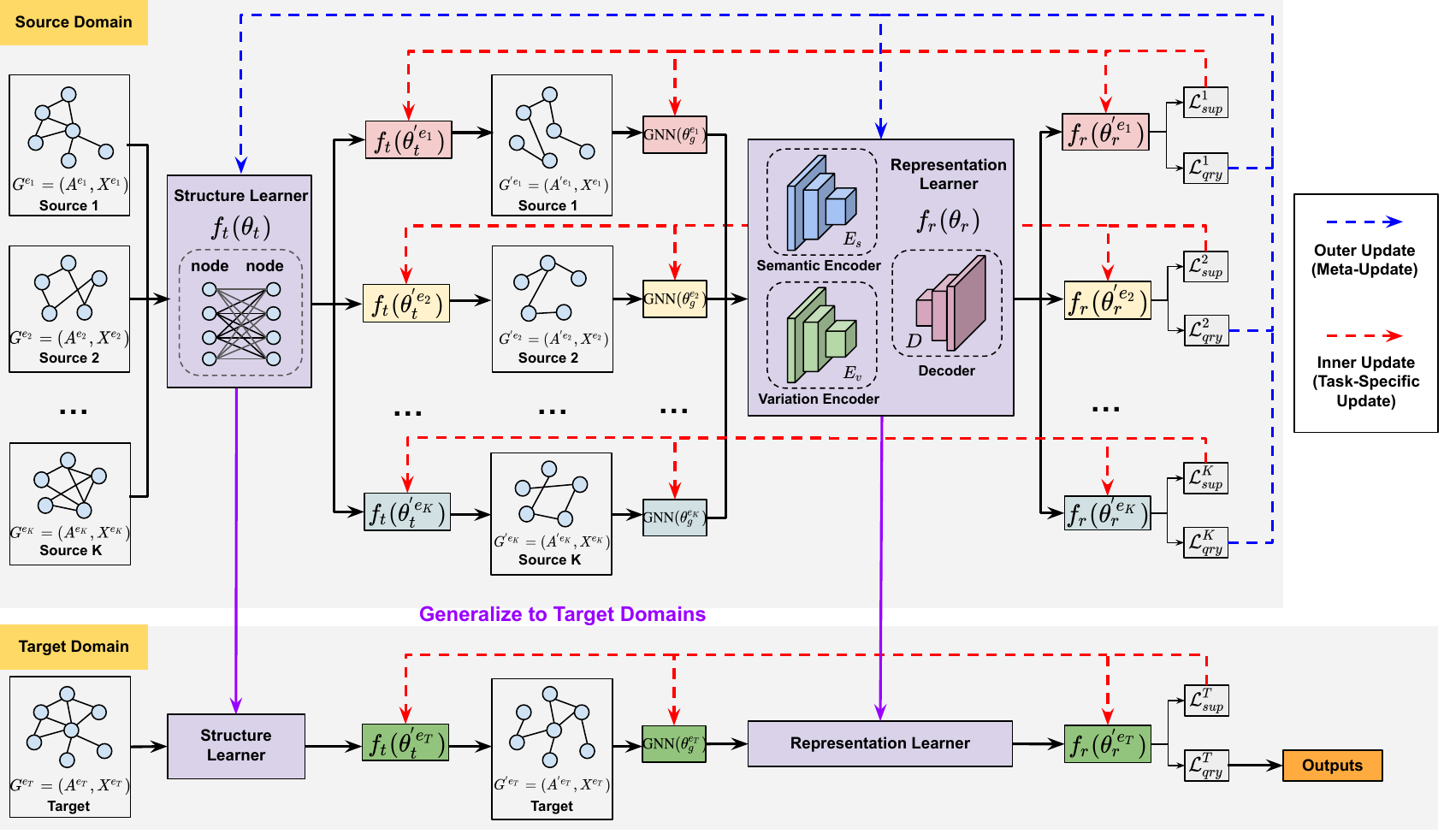}
    \caption{ An overview of \sysname{}. 
    Each source graph is viewed as a task. For each task, the parameters $\{ \boldsymbol{\theta}'_t,\boldsymbol{\theta}'_g,\boldsymbol{\theta}'_r\}$ of the structure learner ($f_t$), GNN, and representation learner ($f_r$) are updated via $\mathcal{L}_{sup}$ during the inner update phase. Subsequently, the query losses $\mathcal{L}_{qry}$ across all tasks are aggregated to update the meta-parameters $\boldsymbol{\theta} =\{ \boldsymbol{\theta}_t, \boldsymbol{\theta}_r\}$ in the outer update phase. To generalize to graphs in the target domain, the learned meta-parameters of the structure learner and the representation learner are further fine-tuned for adaptation. }
    \label{fig:framework}
    % \vspace{-4mm}
\end{figure}

% which can improve the robustness of GNNs and capture the shared structure information across domains;
\textbf{Problem Setup.} 
% Each graph $G^e = (A^e,X^e,\mathbf{y}^e)$.
To address the problem of node-level domain generalization on graphs, where each domain is characterized by variations on both topology structures and node attributes, learning cross-multi-domain shared graph topology and node representation information is essential for capturing transferable knowledge across different domains. 
As shown in Fig.~\ref{fig:framework}, a novel framework \sysname{} is proposed in the context of meta-learning with two key components: a structure learner $f_t:\Theta\times(\mathcal{A}\times\mathcal{X})\rightarrow\mathcal{A}$ parameterized by $\boldsymbol{\theta}_t$ and a representation learner $f_r:\Theta\times\mathbb{R}^d\rightarrow\mathbb{R}^d$ parameterized by $\boldsymbol{\theta}_r$.
The goal of \sysname{} aims to learn a good parameter initialization $\boldsymbol{\theta}=\{\boldsymbol{\theta}_t,\boldsymbol{\theta}_r\}$ across all given source graphs $\{G^{e_i}\}_{i=1}^K, \forall e_i\in\mathcal{E}_s$, such that the learned $\boldsymbol{\theta}$ can be effectively adapted to the target graph $G^{e_T},e_T\in\mathcal{E}_t$, which is inaccessible during training.
% The learned initialization parameters $\boldsymbol{\theta}$ include a structure learner parameter $\boldsymbol{\theta}_{t}$ and a representation learner parameter $\boldsymbol{\theta}_{r}=(\boldsymbol{\theta}_{s}, \boldsymbol{\theta}_{v},\boldsymbol{\theta}_{d})$.
% % We frame our training goal as a bi-level optimization problem:}
% We frame our training goal as follows:
\begin{equation}
% \footnotesize
\begin{aligned}
    &\boldsymbol{\theta} = \arg\min_{\boldsymbol{\theta}} \frac{1}{M}\sum_{i=1}^{M} \ell(\boldsymbol{\theta}^{'e_i},\mathcal{T}^{e_i}_{qry}), \quad \forall e_i \in\mathcal{E}_s\\
    \text{where}\: 
    % &\boldsymbol{\theta}=\{\boldsymbol{\theta}_t,\boldsymbol{\theta}_r\} \:\text{and}\: \boldsymbol{\theta}^{'{e_i}} =  \{\boldsymbol{\theta}_t^{'e_i},\boldsymbol{\theta}_r^{'e_i}\},\\
    &\boldsymbol{\theta}^{'e_i} = \boldsymbol{\theta}-\alpha \nabla \ell (\boldsymbol{\theta},\mathcal{T}_{sup}^{e_i}),\\
    &\mathcal{T}_{sup}^{e_i} = E_s\Big(\boldsymbol{\theta}_s ,\text{GNN}(\boldsymbol{\theta}_g^{e_i}, G^{e_i})\oplus\text{GNN}(\boldsymbol{\theta}_g^{e_i}, G^{'e_i}=(X^{e_i},f_t(\boldsymbol{\theta}_t, G^{e_i}))\Big),
\end{aligned}
\label{problem}
\end{equation}
where $M\leq K$ represents the number of tasks and $\oplus$ is denoted as element-wise addition operation. $\text{GNN}:\Theta\times(\mathcal{A}\times\mathcal{X})\rightarrow\mathbb{R}^d$ is a graphic representation function, parameterized by $\boldsymbol{\theta}_g^{e_i}$ specific to the domain $e_i$. 
The representation learner $f_r$ consists of a semantic encoder $E_s:\Theta\times\mathbb{R}^d\rightarrow\mathbb{R}^s$, a variation encoder  $E_v:\Theta\times\mathbb{R}^d\rightarrow\mathbb{R}^v$, and a decoder $D:\Theta\times\mathbb{R}^{s+v}\rightarrow\mathbb{R}^d$. 
We thus denote $\boldsymbol{\theta}_r$ as consisting of the parameters $\boldsymbol{\theta}_s$, $\boldsymbol{\theta}_v$, and $\boldsymbol{\theta}_d$, respectively, \textit{i.e.,} $\boldsymbol{\theta}_r=\{ \boldsymbol{\theta}_s,\boldsymbol{\theta}_v,\boldsymbol{\theta}_d\}$.
Detailed setting and training of $f_r$ is introduced in Section \ref{sec:RL}.
$\mathcal{T}_{sup}^{e_i}$ and $\mathcal{T}_{qry}^{e_i}$ are support and query sets of the domain $e_i$, which are randomly sampled from the output of the semantic encoder $E_s$.
Inspired by MAML, meta-parameters $\boldsymbol{\theta}=\{\boldsymbol{\theta}_t,\boldsymbol{\theta}_r\}$ and task-specific parameters $\boldsymbol{\theta}^{'e_i}=\{\boldsymbol{\theta}_t^{'e_i},\boldsymbol{\theta}_r^{'e_i}\}$ are updated interchangeably through the bi-level optimization, using query sets $\{\mathcal{T}_{qry}^{e_i}\}_{i=1}^M$ and support sets  $\{\mathcal{T}_{sup}^{e_i}\}_{i=1}^M$, respectively. 
The learned $\boldsymbol{\theta}$ is further fine-tuned to $\boldsymbol{\theta}^{'e_T}$ using $\mathcal{T}^{e_T}_{sup}$ from the target domain. The generation performance is then evaluated on $\boldsymbol{\theta}^{'e_T}$ using $\mathcal{T}^{e_T}_{qry}$.

\section{Methodology}
% We now present the details of our \sysname{} framework.
The primary challenges of \sysname{} involve modeling and capturing generalizable structure patterns across multiple domains, as well as disentangling domain-invariant semantic factors and domain-specific variation factors. 
To this end, the structure learner $f_t$ is devoted to capturing shared structural information across domains while enhancing the comprehensiveness of representations learned by
GNN through mitigating the adverse effects of task-unrelated edges (Sec.~\ref{sec:SL}).
Additionally, the representation learner $f_r$ captures the invariant patterns of the truly predictive properties through the semantic encoder $E_s$ by disentangling semantic and variation factors in node representations based on the causal invariance principle (Sec.~\ref{sec:RL}).
Finally, in Sec.~\ref{sec:MAML}, we integrate two learners within the meta-learning framework to capture transferable knowledge across various domains.
For simplicity, in this section, domain $e_i$ is simplified to $i$.

\subsection{Structure Learner}
\label{sec:SL}
For graph data with both attributes and topologies, how to learn as comprehensive and rich node representation as possible is a problem that has been explored. 
One prevalent method is GNNs, which learns node representations through recursive aggregation of information from neighboring nodes. 
However, based on the model of the message-passing mechanism, small noise propagation to neighboring areas may cause deterioration of the representation quality. Therefore, we optimize GNN by learning high-quality graph structures. 
% Further, we also explore the common structural pattern between cross-domain graphs to improve generalization ability by learning a cross-domain shared structural encoder.
Further, we also explore the common structural pattern between cross-domain graphs to improve generalization ability.
Here, we define a refined graph structure matrix as $A'$ learned by a graph structure learner $f_t$.
The $f_t$ is expected to produce optimal graph structures that can give rise to satisfactory downstream classification performance.

First, we learn an intermediate similarity graph matrix $F$, where $F_{jk}$ denotes the edge weights of node $j$ and $k$.
To fuse attributes and topological information, we use the representation of nodes $\mathbf{r} \in \mathbb{R}^d$ to calculate the weight of edges between nodes:
\begin{equation}
% \footnotesize
        \begin{split}
        F_{jk} = \delta(\mathbf{r}_j\odot \mathbf{\hat{w}}, \mathbf{r}_k\odot \mathbf{\hat{w}})
        \end{split},
    \label{eq.edge_weight}
\end{equation}
where $\mathbf{\hat{w}} \in \mathbb{R}^{m}$ is a weight vector and is trainable, $\delta(\cdot, \cdot)$ is a similarity function that includes simple dot-product and so on.
After obtaining $F$, we generate a novel graph structure $A'$ by sampling from ${A'}_{jk} \sim Bernoulli (F_{jk})$.
Further, we regularize the learned graph structure using sparsity and smoothness constraints:
\begin{equation}
% \footnotesize
        \begin{split}
        \mathcal{B} =-\alpha \sum_{j,k} {A}'_{j,k} || \mathbf{r}_{j}-\mathbf{r}_{j} ||_2^2 - \beta ||{A}'||_{0}
        \end{split},
        \label{eq.regularity}
\end{equation}
where the $\alpha$ and $\beta$ are hyperparameters controlling different modules' importance.
We adopt the policy gradient optimization method for the non-differentiable problem of sampling $A'$.
We define the probability for sampling as follows:
\begin{equation}
% \footnotesize
    \Phi(A')= \Pi_{j,k}\left(A'_{jk}F_{jk} + (1- A'_{jk})(1-F_{jk})\right).
    \label{eq.sample_probability}
\end{equation}
Then we independently sample $H$ times to obtain $\{A'\}_{h=1}^{H}$ and $\{\Phi(A')\}_{h=1}^{H}$.
We define the regularization $\mathcal{B}$ in Eq.~\eqref{eq.regularity} as the reward function, then we optimize the $\boldsymbol{\theta}_{t}$ using REINFORCE~\cite{zhang2021sample} algorithm with the gradient:
\begin{equation}
  % \footnotesize  
  \nabla_{\boldsymbol{\theta}_{t}}\mathcal{L}^{reg}= -\nabla_{\boldsymbol{\theta}_{t}}\frac{1}{H} \sum_{h=1}^{H} \log{\Phi(A'_{h})(\mathcal{B}(A'_h)-\bar{\mathcal{B}})},
    \label{reg_opt}
\end{equation}
where $\bar{\mathcal{B}}$ is the mean value,
serving as a baseline function.
It operates by averaging the regularization rewards $\mathcal{B}(A')$  across a single feed-forward computation, thereby aiding in the reduction of variance throughout the training of the policy gradient.
% Nevertheless, adopting such an approach gives rise to quadratic algorithmic complexity ($O(N^2)$).
% Inspired by \cite{zhao2023graphglow}, we choose $H$ nodes as hubs and transform the $N \times N$ graph to the product of $N \times H$ node-hub bipartite graph ${H}$ and $H \times N$ hub-node graph bipartite ${H}^T$, representing the node-hub relationship.
% $H$ is a hyperparameter much smaller than $N$ and enables a significant reduction in computational complexity to $O(NP)$.
% In this way, we can compute a node-hub similarity matrix $\hat{H}$ based on (\ref{eq.edge_weight}).
% Then we sample from each $\hat{H}_{ij} \sim Bernoulli (\hat{H}_{ij})$ to obtain ${H}$.
% When we compute the Eq.~\eqref{eq.regularity}, the computation still requires $O(N^2)$. 
% We apply regularization to the $H \times H$ pivot-pivot adjacency matrix ${B} = H^TH$ to alleviate overhead, serving as a proxy regularization:
% \begin{equation}
%         \begin{split}
%         -\alpha \sum_{i.j}\hat{A}_{i,j} || \mathbf{r}_{i}-\mathbf{r}_{j} ||_2^2 - \beta ||\hat{B}||_{0}
%         \approx -\alpha \sum_{i.j}\hat{B}_{i,j} || \mathbf{r'}_{i}-\mathbf{r'}_{j} ||_2^2 - \beta ||\hat{H}||_{0}
%         \end{split}
%         \label{eq.reg_opt}
% \end{equation}
% where $\mathbf{r}'$ represents the feature of the $i$-th hub node.

After obtaining $A'$ by structure learner $f_t$, we recursively propagate features along the latent graph $A'$ and original adjacent matrix $A$ to update node representations by the GNN network parameterized by $\mathbf{\theta}_g$.
This process can be formulated as:
\begin{equation}
% \footnotesize
     R = \lambda \text{GNN}(R, A) + (1-\lambda)\text{GNN}(R, A'),
    \label{eq.representation}
\end{equation}
where $\lambda$ denotes the weight coefficient, $R=\{ \mathbf{r}_j\}_{j=1}^{|\mathcal{V}|} \in \mathbb{R}^{|\mathcal{V} \times d}$ where $\mathbf{r} \in \mathbb{R}^d$ is the node representation.

% Then $\mathcal{R}_k = \text{GNN}(A_k,X_k,A'_k;\mathbf{w}_k)$, where $\mathcal{R}_k =\{\mathbf{r}_v\}_{v \in {G}_k}$.

\subsection{Representation Learner}
\label{sec:RL}
%\label{Semantic Learner}
% \begin{figure}[t]
%     \centering
%     \includegraphics[width=\linewidth]{figures/semantic.pdf}
%     \caption{The representation learner}
%     \label{fig:semantic}
% \end{figure}

% After obtaining the representation of each node,
Despite GNNs having the capability to extract abstract representations for predictions, the representation may unconsciously mix up semantic factors $\mathbf{s}$ and variation factors $\mathbf{v}$ due to a correlation between them.
So the model still relies on the domain-specific variation factors $\mathbf{v}$ for prediction via this correlation.
However, this correlation may change drastically in a new domain, making the effect from $\mathbf{v}$ misleading.
So we assume the representation of each node is disentangled into two factors: a domain-invariant semantic factor $\mathbf{s}$ determining the label and a domain-specific variation factor $\mathbf{v}$ independent of
labels.
% The goal of the representation learner is to capture two latent factors $\mathbf{s}$ and $\mathbf{v}$.
% In this section, we use lowercase letters to represent all variables.
$p(\mathbf{r}|\mathbf{s},\mathbf{v})$ and $p(\mathbf{y}|\mathbf{s})$ are invariant across domains, and the change of prior $p(\mathbf{s},\mathbf{v})$ is the only source of domain change.
Based on the above causal generative principle, we develop the representation learner based on variational Bayes~\cite{kingma2013auto,jordan1999introduction}.
Here, the representation of the node learned by GNN $\mathbf{r}$ and the label $\mathbf{y}$ are accessible variables and we have supervised data from the underlying representation $p^{*}(\mathbf{r},\mathbf{y})$ in the source domain.
The log marginal likelihood of the $\mathbf{r}$ and $\mathbf{y}$ is as follows:

\begin{equation}
% \footnotesize
    \log p(\mathbf{r},\mathbf{y}) = \log \int \int p(\mathbf{s},\mathbf{v},\mathbf{r},\mathbf{y}) d\mathbf{s}d\mathbf{v}
    \label{eq.likelihood},
\end{equation}
where $p(\mathbf{s},\mathbf{v},\mathbf{r},\mathbf{y}) := p(\mathbf{s},\mathbf{v})p(\mathbf{r}|\mathbf{s},\mathbf{v})p(\mathbf{y}|\mathbf{s})$.
By maximizing the likelihood in Eq.~\eqref{eq.likelihood}, $p(\mathbf{r},\mathbf{y})$ will match the $p^{*}(\mathbf{r},\mathbf{y})$.
However, the direct optimization of Eq.~\eqref{eq.likelihood} is intractable, so we utilize the variational inference~\cite{jordan1999introduction} to approximate the marginal likelihood.
We introduce a tractable distribution $q(\mathbf{s}, \mathbf{v}|\mathbf{r},\mathbf{y})$ and construct the variational objective as follows:
\begin{equation}
% \footnotesize
        \begin{split}
          \log p(\mathbf{r},\mathbf{y}) \geqslant \mathbb{E}_{q(\mathbf{s},\mathbf{v}|\mathbf{r},\mathbf{y})} \left [\log \frac{p(\mathbf{s},\mathbf{v},\mathbf{r},\mathbf{y})}{q(\mathbf{s},\mathbf{v}|\mathbf{r},\mathbf{y})} \right] 
          =: \mathcal{L}_{q_{\mathbf{s},\mathbf{v}|\mathbf{r},\mathbf{y}}}(\mathbf{r},\mathbf{y}),
        \end{split}
        \label{eq.ELBO}
\end{equation}
% Then we obtain $p(r,y) := \int p(s,v,r,y) dsdv$ where $p(s,v,r,y) := p(s,v)p(r|s,v)p(y|s)$, which is hard to estimate.
% A common and effective approach to let the model $p$ match the data distribution $p^*(r,y)$ is maximizing likelihood $\mathbb{E}_{p^*(r,y)}[\log p(r,y)]$, where $p := \langle p(s,v), p(r| s,v), p(y|s) \rangle$.
% However, it is hard to estimate and optimize because $p(r,y) := \int p(s,v,r,y) dsdv$ where $p(s,v,r,y) := p(s,v)p(r|s,v)p(y|s)$ is intractable.
% To circumvent the difficulty, we use variational expectation-maximization (variational EM) to introduce a tractable distribution $q(s, v|r, y)$.
% According to Jensen’s inequality and the concavity of the log function, a lower bound of the likelihood function can be derived:
where $\mathcal{L}_{p,q_{\mathbf{s},\mathbf{v}|\mathbf{r},\mathbf{y}}}(\mathbf{r},\mathbf{y})$ is called Evidence Lower BOund (ELBO).
% The intricacy inherent in supervised learning lies in the persistent difficulty of prediction. 
Unfortunately, the introduced model $q(\mathbf{s}, \mathbf{v}|\mathbf{r}, \mathbf{y})$ fails to facilitate the estimation of $p(\mathbf{y}|\mathbf{r})$.
To alleviate this problem, we introduce an auxiliary model $q(\mathbf{s}, \mathbf{v}, \mathbf{y}|\mathbf{r})$ to target $p(\mathbf{s}, \mathbf{v}, \mathbf{y}|\mathbf{r})$, which enables the straightforward sampling of $\mathbf{y}$ given $\mathbf{r}$ for prediction. 
Meanwhile, $q(\mathbf{s},\mathbf{v}, \mathbf{y}|\mathbf{r}) = q(\mathbf{s}, \mathbf{v}|\mathbf{r}, \mathbf{y})q(\mathbf{y}|\mathbf{r})$ means it can help learning inference model $q(\mathbf{s}, \mathbf{v}|\mathbf{r}, \mathbf{y})$, where $q(\mathbf{y}|\mathbf{r}) := \int q(\mathbf{s}, \mathbf{v}, \mathbf{y}|\mathbf{r})d\mathbf{s}d\mathbf{v}$.
% Then, The ELBO objective $\mathbb{E}_{p^*(r,y)}[\log p(r,y)]$ following:
% \begin{equation}
%         \footnotesize
%         \begin{split}
%         \mathbb{E}_{p^*(r)} \mathbb{E}_{p^*(y|r)}\left[\log q(y|r)\right] + \mathbb{E}_{p^*(r)} \mathbb{E}_{q(s,v,y|r)} \left[\frac{p^*(y|s)}{q(y|s)} \log \frac{p(s,v,r,y)}{q(s,v,y|r)} \right].
%         \end{split}
%         \label{eq.ELBO_1}
%     \end{equation}
% In Eq.~\eqref{eq.ELBO_1}, the first term is the negative of the standard cross entropy (CE) loss which drives $q(y|r)$ 
% towards $p^{*}(y|r)$, and once this is achieved, the second term becomes the expected ELBO $\mathbb{E}_{p^*(r)}[\mathcal{L}_{p,q_{s,v,y|r}}(r)]$ which drives $q(s, v, y|r)$ towards $p(s, v, y|r)$ and $p(r)$ towards $p^*(r)$.
Due to $p(\mathbf{s}, \mathbf{v}, \mathbf{y}|\mathbf{r})$ can be factorized as $p(\mathbf{s}, \mathbf{v}|\mathbf{r})p(\mathbf{y}|\mathbf{s})$.
Thus, we can instead introduce a lighter inference $q(\mathbf{s},\mathbf{v}|\mathbf{r})$ for the minimally intractable component $p(\mathbf{s}, \mathbf{v}|\mathbf{r})$ and use $ q(\mathbf{s}, \mathbf{v}|\mathbf{r})p(\mathbf{y}|\mathbf{s})$ as an approximation to $q(\mathbf{s}, \mathbf{v}, \mathbf{y}|\mathbf{r})$.
This turns the objective Eq.~\eqref{eq.ELBO} to:
% \begin{equation}
% \footnotesize
%         \begin{split}
%         \max_{p,q_{s,v|r}} \mathbb{E}_{p^*(r,y)}\left[ \mathbb{E}_{q(s,v|r)}\left[ \log p(y|s)\right] + \frac{1}{p} \mathbb{E}_{q(s,v|r)}\left[p(y|s) \log \frac{p(s,v)}{q(s,v|r)}  + \log p(r|s,v)\right]\right],
%         \end{split}
%         \label{eq.ELBO_loss}
% \end{equation}
\begin{equation}
% \footnotesize
        \begin{split}
         \log p(\mathbf{r},\mathbf{y}) 
         &\geqslant
         \frac{1}{q(\mathbf{y}|\mathbf{r})} \left[ \mathbb{E}_{q(\mathbf{s},\mathbf{v}|\mathbf{r})}\left [p(\mathbf{y}|\mathbf{s}) \log q(\mathbf{y} | \mathbf{r})\right] \right]\\
         &+\frac{1}{q(\mathbf{y}|\mathbf{r})} \left[\mathbb{E}_{q(\mathbf{s},\mathbf{v}|\mathbf{r})}\left[p(\mathbf{y}|\mathbf{s}) \log p(\mathbf{r} | \mathbf{s},\mathbf{v})\right] \right]\\
         &+  \frac{1}{q(\mathbf{y}|\mathbf{r})} \left[\mathbb{E}_{q(\mathbf{s},\mathbf{v}|\mathbf{r})}\left[p(\mathbf{y}|\mathbf{s}) \log \frac{p(\mathbf{s}, \mathbf{v})}{q(\mathbf{s},\mathbf{v}|\mathbf{r})} \right] \right]\\
          % \mathbb{E}_{q(\mathbf{s},\mathbf{v}|\mathbf{r})}\left[ \log p(\mathbf{y}|\mathbf{s})\right] \\
          %  & + \frac{\mathbb{E}_{q(\mathbf{s},\mathbf{v}|\mathbf{r})}\left[p(\mathbf{y}|\mathbf{s}) \log p(r|\mathbf{s},\mathbf{v})\right]}{\mathbb{E}_{q(\mathbf{s},\mathbf{v}|\mathbf{r})} \left[p(\mathbf{y}|\mathbf{s}) \right]} \\
          % & + \frac{\mathbb{E}_{q(\mathbf{s},\mathbf{v}|\mathbf{r})}\left[p(\mathbf{y}|\mathbf{s}) \log \frac{p(\mathbf{s},\mathbf{v})}{q(\mathbf{s},\mathbf{v}|\mathbf{r})} \right]}{\mathbb{E}_{q(\mathbf{s},\mathbf{v}|\mathbf{r})} \left[p(\mathbf{y}|\mathbf{s}) \right]} \\
          &=: \mathcal{L}^{ELBO},
        \end{split}
        \label{eq.ELBO_loss}
\end{equation}
where $q(\mathbf{y}|\mathbf{r}) = \mathbb{E}_{q(\mathbf{s},\mathbf{v}|\mathbf{r})} \left [ p(\mathbf{y}|\mathbf{s})\right]$.
The $\mathcal{L}^{ELBO}$ in Eq.~\eqref{eq.ELBO_loss} consists of three components. The first term is the negative of the standard cross entropy (CE) loss and $p(\mathbf{y}|\mathbf{s})$ gives the ability to model to predict the target label.
The second term encourages the latent representation $\mathbf{s}$ and $\mathbf{v}$ to preserve the salient information of $\mathbf{r}$ by reconstruction.
The third and fourth term drives the variational posteriors $q(\mathbf{s},\mathbf{v}|\mathbf{r})$ towards its priors.
By maximizing the $\mathcal{L}^{ELBO}$ in Eq.~\eqref{eq.ELBO_loss}, it becomes feasible to deduce the parameters of distribution over the joint latent variables $\mathbf{s}$ and $\mathbf{v}$.
The Monte Carlo method can be used to estimate expectations~\cite{kingma2013auto}. 
The derivation of Eq.~\eqref{eq.ELBO_loss} is provided in Appendix~\ref{model_details}.

\textbf{\sysname{}-ind}. 
To improve the generalization of the model, we consider another case where $\mathbf{s}$ and $\mathbf{v}$ are independent, \textit{i.e.,} $p^{\perp}(\mathbf{s},\mathbf{v}) = p(\mathbf{s})p(\mathbf{v})$. 
% This implies that the $\mathbf{s}$ and $\mathbf{v}$ is disentangled.
Formally, the distribution $p^{\perp}(\mathbf{s},\mathbf{v})$ exhibits a higher entropy compared to $p(\mathbf{s},\mathbf{v})$, which diminishes specific information of the source domain and promotes dependence on causal mechanisms for enhanced generalization.
Under the conditional independence assumption, the $p(\mathbf{s}, \mathbf{v}))$ can be turned to $p(\mathbf{s})p(\mathbf{v})$ and  $q(\mathbf{s},\mathbf{v}|\mathbf{r})$ can be turned to $q(\mathbf{s}|\mathbf{r})q(\mathbf{v}|\mathbf{r})$ in Eq.~\eqref{eq.ELBO_loss}, which is denoted as $\mathcal{L}^{ELBO}_{ind}$.
% \begin{equation}
% \footnotesize
%         \begin{split}
%          \log p(\mathbf{r},\mathbf{y}) 
%          &\geqslant
%          \frac{1}{q(\mathbf{y}|\mathbf{r})} \left[ \mathbb{E}_{q(\mathbf{s},\mathbf{v}|\mathbf{r})}\left [p(\mathbf{y}|\mathbf{s}) \log q(\mathbf{y} | \mathbf{r})\right] \right]\\
%          &+\frac{1}{q(\mathbf{y}|\mathbf{r})} \left[\mathbb{E}_{q(\mathbf{s},\mathbf{v}|\mathbf{r})}\left[p(\mathbf{y}|\mathbf{s}) \log p(\mathbf{r} | \mathbf{s},\mathbf{v})\right] \right]\\
%          &+  \frac{1}{q(\mathbf{y}|\mathbf{r})} \left[\mathbb{E}_{q(\mathbf{s},\mathbf{v}|\mathbf{r})}\left[p(\mathbf{y}|\mathbf{s}) \log \frac{p(\mathbf{s})p(\mathbf{v})}{q(\mathbf{s}|\mathbf{r})q(\mathbf{v}|\mathbf{r})} \right] \right]\\
%           &=: \mathcal{L}^{ELBO}_{ind}.
%         \end{split}
%         \label{eq.ELBO_loss_ind}
% \end{equation}
% \begin{equation}
% \footnotesize
%         \begin{split}
%          \log p(\mathbf{r},\mathbf{y}) &\geqslant
%           \mathbb{E}_{q(\mathbf{s},\mathbf{v}|\mathbf{r})}\left[ \log p(\mathbf{y}|\mathbf{s})\right] \\
%           & +\mathbb{E}_{q(\mathbf{s},\mathbf{v}|\mathbf{r})}\left[ \log p(\mathbf{r}|\mathbf{s},\mathbf{v})\right] \\
%           & - KL\left(q(\mathbf{s}|\mathbf{r},\mathbf{y}) || p(\mathbf{s})\right) \\
%           & -KL\left(q(\mathbf{v}|\mathbf{r},\mathbf{y}) || p(\mathbf{v}) \right) \\
%           & =: \mathcal{L}^{ELBO}
%         \end{split}
%         \label{eq.ELBO_loss_ind}
% \end{equation}

\subsubsection{Model in Detail.}
We show the distributional assumption in \sysname{}.
For the prior $p(\mathbf{s},\mathbf{v})$ , we use a multivariate Gaussian distribution $p(\mathbf{s},\mathbf{v}) = \mathcal{N} \left( (\mathbf{s},\mathbf{v}) | (\mu_{s}, \mu_{u}), \sum \right)$, where $\mu_{s}$ and $\mu_{v}$ are both zero vectors and
$\sum=\begin{pmatrix}
    \sum_{ss} &\sum_{sv} \\
    \sum_{vs} &\sum_{vv}
\end{pmatrix}$
is parameterized by its Cholesky decomposition~\cite{higham1990analysis}.
For semantic encoder $E_{s}$ and variation encoder $E_{v}$, we formulate their variational posterior distribution as Gaussian distribution with diagonal covariance structure, parameterized by neural network.
For the decoder $D$, we adopt the Gaussian distribution $q(\mathbf{s},\mathbf{v}|\mathbf{r}) = \mathcal{N}(\mathbf{s},\mathbf{v}| \mu_{r}, \sigma_{r}^2I)$, where $\mu_{r}$ and $\sigma_{r}$ are given by the mapping during reconstruct process.
The semantic factor $\mathbf{s}=E_{s}(\mathbf{r})$ is used to predict node label $y$, \textit{i.e.,} $\hat{y} = \hat{f}(\mathbf{s})$, where $\hat{f}$ is a classifier.
The variation factor $\mathbf{v}=E_{v}(\mathbf{r})$ is independent of label.
\sysname{} uses both $\mathbf{s}$ and $\mathbf{v}$ to reconstruct representation $\mathbf{r}$, \textit{i.e.,} $\mathbf{r}=D(\mathbf{s},\mathbf{v})$.
For the prior $p(\mathbf{s})$ and $p(\mathbf{v})$ in \sysname{}-ind, we adopt standard Gaussian $\mathcal{N}(\mathbf{s};0,I)$ and $\mathcal{N}(\mathbf{v};0,I)$, respectively.
% Due to the limited space, we show the other details of distributional assumption in the Appendix~\ref{setting}).

%\subsubsection{Instantiation for $q(s,v|r)$}
% The variational distribution aims at learning the conditional distribution that generates a suitable semantic factor and a variation factor based on the latent representation of an input graph.
% We define a semantic encoder $E_{\boldsymbol{\theta_s^*}}^{s}$, \textit{i.e.,} $s = E_{\boldsymbol{\theta_s^*}}^{s}(r)$ and a variation encoder $E_{\boldsymbol{\theta_v^*}}^{v}$, \textit{i.e.,} $v = E_{\boldsymbol{\theta_v^*}}^{v}(r)$.
% Based on the above description, the semantic factor $s$ is what determines the label $Y$, so we want to get the following objective function:

% \begin{equation}
%         \begin{split}
%         \mathcal{L}^{ce} =l(f_{\boldsymbol{\theta}}(s), Y)
%         \end{split}
%         \label{eq.loss_ce}
% \end{equation}
% where $l$ is the cross entropy function.

% \subsubsection{Instantiation for $p(r|s,v)$}
% We define a decoder $D$, which is used to reconstruct the representation $r'$, \textit{i.e.,} $r' = D(s,v)$.
% The objective function can be
% formulated as the following:
% \begin{equation}
%         \begin{split}
%         \mathcal{L}^{rec} = \parallel D(s,v), r \parallel ^2
%         \end{split}
%         \label{eq.loss_rec}
% \end{equation}

% $\lambda_{ce}$, $\lambda_{rec}$, and $\lambda_{elbo}$ are the weight coefficient of each module.

\subsection{Meta-Learning}
\label{sec:MAML}
% The goal of \sysname{} is to learn the initialization parameters of the structure learner with $\boldsymbol{\theta}_t$, representation learner with $\boldsymbol{\theta}_s$, $\boldsymbol{\theta}_v$ and $\boldsymbol{\theta}_d$, which can be directly applied to target graph inference with simple fine-tuning.
To learn a good parameter initialization of $\boldsymbol{\theta}=\{\boldsymbol{\theta}_t,\boldsymbol{\theta}_r\}$ across all given source graphs, we use the MAML~\cite{finn2017model} framework to integrate the structure learner with the representation learner. 
The objective function is:
\begin{equation}
% \footnotesize
        \begin{split}
        \mathcal{L} =  -\mathcal{L}^{ELBO} + \lambda_{r}\mathcal{L}^{reg},
        \end{split}
        \label{eq.loss_total}
\end{equation}
where $\lambda_{r}$ is the weight coefficient of  regularization loss.
We randomly sampled the support set $\mathcal{T}_{sup}^{i}$ and the query set $\mathcal{T}_{qry}^{i}$ from the output of the semantic encoder $E_s$  for each task $i$.

For the inner update, first, compute $\mathcal{L}_{sup}^i$ on $\mathcal{T}_{sup}^{i}$, and then update its parameters $\boldsymbol{\theta}'^i$ iteratively with $\eta$ loops as follows:
% \begin{equation}
% \footnotesize
%         \begin{split}
%         \mathcal{L} =  -\mathcal{L}^{ELBO} + \lambda_{r}\mathcal{L}^{reg},
%         \end{split}
%         \label{eq.loss_total}
% \end{equation}
\begin{equation}
% \footnotesize
        \begin{split}
        \boldsymbol{\theta}'^i = \boldsymbol{\theta} - l_{in} \nabla_{\boldsymbol{\theta}}  \mathcal{L}^i_{sup},
        \end{split}
        \label{eq.inner_loop}
\end{equation}
where $l_{in}$ is the inner learning rate and $\boldsymbol{\theta}'^i = \{\boldsymbol{\theta}'^i_t, \boldsymbol{\theta}^i_g,\boldsymbol{\theta}'^i_s,\boldsymbol{\theta}'^i_v, \boldsymbol{\theta}'^i_d\}$.

% For the inner update, the process begins with learning the representation $\mathbf{r}$ through the structure learner and GNN.
% Subsequently, semantic factor $\mathbf{s}$ and variation factor $\mathbf{v}$ are captured for each domain by the representation learner.
% The semantic factor encoded by $E_s$ can be used to predictions of input graph $G$, \textit{i.e.,} $\hat{\mathbf{y}} = \hat{f}(\mathbf{s})$, where $\hat{f}$ is a classifier for each domain.

% It is worth noting that in our article, each domain corresponds to a domain-specific  $E^s_{\boldsymbol{\theta}_s}$ and $E^v_{\boldsymbol{\theta}_v}$.
% Let $E^s_{\boldsymbol{\theta_s}}$ be a trained semantic encoder parameterized by weights $\boldsymbol{\theta_s}$ and a variation encoder $E^v_{\boldsymbol{\theta_v}}$ parameterized by weights $\boldsymbol{\theta_v}$.

For the outer update, we apply the parameters $\boldsymbol{\theta}'^i$ that have been iteratively updated several times in the inner loop to its query set $\mathcal{T}_{qry}^{i}$ to calculate the $\mathcal{L}^i_{qry}$.
Then update $\boldsymbol{\theta}$ as follows:
\begin{equation}
% \footnotesize
        \begin{split}
   \boldsymbol{\theta} = \boldsymbol{\theta} - l_{out} \nabla_{\boldsymbol{\theta}} \frac{1}{M}\sum_{i=1}^{M} \mathcal{L}^i_{qry},
        \end{split}
        \label{eq.outer_loop}
\end{equation}
where $l_{out}$ is the learning rate of the outer loop and $\boldsymbol{\theta}=\{\boldsymbol{\theta}_t, \boldsymbol{\theta}_r\}$. 
The process of our framework is in Algorithm~\ref{whole_algorithm}.
The complexity analysis of the algorithm is presented in Appendix~\ref{complexity}.

\begin{algorithm}[t!]
% \small
    \caption{\textbf{The procedure of \sysname{}}} \label{whole_algorithm}
    \begin{algorithmic}[1]
    \STATE \textbf{Input:} observed source graphs $\{G^i\}_{i=1}^M$, learning rates $l_{in}$, $\l_{out}$
    \STATE Select $M$ source graphs as tasks $\{\mathcal{T}^i\}_{i=1}^M$
    \STATE Initialize $\boldsymbol{\theta}_t$, $\boldsymbol{\theta}_s$, $\boldsymbol{\theta}_v$, $\boldsymbol{\theta}_d$ and
    $\{\boldsymbol{\theta}_g^i\}_{i=1}^M$
    \STATE \textbf{While} not done \textbf{do}: 
    \STATE \qquad \textbf{For} each task $\mathcal{T}^i$ \textbf{do}:
    \STATE \qquad \qquad  $R_0^i =X^i$
    \STATE \qquad \qquad  Compute $F^i$ using Eq.~\eqref{eq.edge_weight}
    \STATE \qquad  \qquad Sample $H$ times over $F^i$ to obtain  $A'^i$
    \STATE \qquad \qquad 
    Compute $R^i$ using Eq.~\eqref{eq.representation}
    % \STATE \qquad \qquad  $\{ \mathbf{r}_{sup}, \mathbf{y}_{sup} \}$, $\{ \mathbf{r}_{qry}, \mathbf{y}_{qry} \}$
    \STATE \qquad \qquad $\boldsymbol{\theta}'^i_{t_{0}} = \boldsymbol{\theta}_t$, $\boldsymbol{\theta}'^i_{s_{0}} = \boldsymbol{\theta}_s$, $\boldsymbol{\theta}'^i_{v_{0}} = \boldsymbol{\theta}_v$, $\boldsymbol{\theta}'^i_{d_{0}} = \boldsymbol{\theta}_d$,$\boldsymbol{\theta}^i_{g_{0}} = \boldsymbol{\theta}_g^i$
    \STATE \qquad \qquad Sample  $\mathcal{T}_{qry}^{i}$ and $\mathcal{T}_{sup}^{i}$
    \STATE \qquad \qquad \textbf{For} n in $1, \dots, \eta$ \textbf{do}:
     \STATE \qquad \qquad \qquad  Compute $\mathcal{L}_{sup}^i$ on $\mathcal{T}_{sup}^{i}$ via Eq.~\eqref{eq.loss_total}
    \STATE \qquad \qquad \qquad
     $\boldsymbol{\theta}'^i_{t_{n}} = \boldsymbol{\theta}'^i_{t_{{n-1}}} - l_{in}\nabla \mathcal{L}_{sup}^i$
     \STATE \qquad \qquad \qquad
     $\boldsymbol{\theta}'^i_{s_{n}} = \boldsymbol{\theta}'^i_{s_{{n-1}}} - l_{in} \nabla \mathcal{L}_{sup}^i$
      \STATE \qquad \qquad \qquad
     $\boldsymbol{\theta}'^i_{v_{n}} = \boldsymbol{\theta}'^i_{v_{{n-1}}} - l_{in} \nabla \mathcal{L}_{sup}^i$
     \STATE \qquad \qquad \qquad
     $\boldsymbol{\theta}'^i_{d_{n}} = \boldsymbol{\theta}'^i_{d_{{n-1}}} - l_{in} \nabla \mathcal{L}_{sup}^i$
      \STATE \qquad \qquad \qquad
     $\boldsymbol{\theta}_{g_n}^i = \boldsymbol{\theta}^i_{g_{{n-1}}} - l_{in} \nabla \mathcal{L}_{sup}^i$
     \STATE \qquad \qquad \qquad Compute $\mathcal{L}_{qry}^{i,n}$ on $\mathcal{T}_{qry}^{i}$ via Eq.~\eqref{eq.loss_total}
     \STATE \qquad \qquad \textbf{End}
     \STATE \qquad \qquad $\mathcal{L}_{qry}^{i} = \mathcal{L}_{qry}^{i,\eta}$
     \STATE \qquad \textbf{End}
    \STATE \qquad Update $\boldsymbol{\theta}_t \leftarrow \boldsymbol{\theta}_t - l_{out} \nabla_{\boldsymbol{\theta}_t} \frac{1}{M} \sum_{i=1}^M \mathcal{L}^i_{qry}$
    \STATE \qquad Update $\boldsymbol{\theta}_s \leftarrow \boldsymbol{\theta}_s - l_{out} \nabla_{\boldsymbol{\theta}_s} \frac{1}{M} \sum_{i=1}^M \mathcal{L}_{qry}^{i} $
    \STATE \qquad Update $\boldsymbol{\theta}_v \leftarrow \boldsymbol{\theta}_v - l_{out} \nabla_{\boldsymbol{\theta}_v} \frac{1}{M} \sum_{i=1}^M \mathcal{L}_{qry}^{i}$
    \STATE \qquad Update $\boldsymbol{\theta}_d \leftarrow \boldsymbol{\theta}_d - l_{out} \nabla_{\boldsymbol{\theta}_d} \frac{1}{M} \sum_{i=1}^M \mathcal{L}_{qry}^{i}$
    \STATE \textbf{End} while
    \STATE \textbf{Output:} trained initialization parameters $\boldsymbol{\theta}_t$, $\boldsymbol{\theta}_s$, $\boldsymbol{\theta}_v$ and $\boldsymbol{\theta}_d$
    \end{algorithmic}
\end{algorithm}

\section{Theoretical Analysis}
This section presents a theoretical analysis of the boundary guarantee for domain generalization errors in the meta-learning framework that integrates the structural and representation learner.

\begin{theorem}[Upper bound: accuracy]\label{theorem-1}
Define the expected error of $\hat{f}$ in representation space as $\epsilon_{Acc}(\hat{f}) = \mathbb{E}[\mathcal{L}(\hat{f}\circ g(A,X), Y)]$.
For any $\hat{f}: \mathbb{R}^s \rightarrow \mathbb{R}$, any representation mapping $g: \mathcal{A}\times\mathcal{X}\to \mathbb{R}^s$, and any loss function $\mathcal{L}:\mathbb{R}\times\mathbb{R}\to \mathbb{R}$ that is upper bounded by $\pi_u$, the expected error of $\hat{f} \circ g:\mathcal{A} \times \mathcal{X} \rightarrow \mathbb{R}$ at any target domain $e_T \in \mathcal{E}_t$ is upper bounded:
\begin{equation}
% \footnotesize
\begin{split}
&\epsilon_{\texttt{Acc}}^{e_T}\left(\hat{f} \circ g\right) \leq \frac{1}{K}\sum_{i=1}^{K} \epsilon_{\texttt{Acc}}^{e_{i}}\left(\hat{f} \circ g\right) + \sqrt{2\mathbb{E}_{{y\sim \mathbb{P}^{e_{i,j}}_Y}} \left[ {d}_{JS} \left(\mathbb{P}^{e_{i}}_{S|Y}, \mathbb{P}^{e_{j}}_{S|Y} \right)^2 \right] } \\
&+ \sqrt{2}\pi_u \underset{i \in [K]}{\max} d_{JS}\left(\mathbb{P}^{e_{T}}_{A,X,Y}, \mathbb{P}^{e_{i}}_{A,X,Y}\right) + \sqrt{2} \pi_u \underset{i,j \in [K]}{\max} d_{JS}\left(\mathbb{P}^{e_{i}}_Y, \mathbb{P}^{e_{j}}_Y\right).
\end{split}
\label{eq:upper_acc}
\end{equation}
\end{theorem}
For simplicity, we omit parameters $\boldsymbol{\theta}$ and its domain $\Theta$. In this paper, the $g$ function is $g=f_t\circ\text{GNN}\circ E_s$. We adopt Jensen-Shannon (JS) distance \cite{endres2003new} denoted as $d_{JS}$ to quantify the dissimilarity between two distributions. 
% Readers can refer to the details in Appendix~\ref{proof}.
% The right-hand side of Eq.(\ref{eq:upper_acc}) can be only optimized for the first two terms, and the data itself determines the other two terms and can be considered as constants. 
% The specific optimization of Inequality.~\eqref{eq:upper_acc} is discussed in Section.~\ref{app:challenge}.

\begin{theorem}[Lower bound: accuracy]\label{theorem-2}
Suppose $\mathcal{L}(\hat{f} \circ 
 g(A,X),Y)$ is lower bounded by $\pi_c$ when $\hat{f}\circ 
 g(A,X) \neq Y$, and is 0 when $\hat{f}\circ 
 g(A,X)= Y$. 
Let $\xi$ denote the number of labels, if $d_{JS}(\mathbb{P}^{e_{i}}_{Y},\mathbb{P}^{e_{T}}_{Y})\geq  d_{JS}(\mathbb{P}^{e_{i}}_{S},\mathbb{P}^{e_{T}}_{S})$, the expected error of $\hat{f}$ at source and target domains is lower bounded:
\begin{equation}
% \footnotesize
\label{eq:lower}
    \begin{split}
         \frac{1}{K}\sum_{i=1}^{K}\epsilon_{\texttt{Acc}}^{e_{i}}(\hat{f} \circ g)+\epsilon_{\texttt{Acc}}^{e_T}(\hat{f} \circ g)
    \geq  \frac{\pi_c}{4\xi K}\sum_{i=1}^{K}\Big(d_{JS}(\mathbb{P}^{e_{i}}_{Y},\mathbb{P}^{e_T}_{Y})- d_{JS}(\mathbb{P}^{e_{i}}_{A,X},\mathbb{P}^{e_T}_{A,X}) \Big)^4.
    \end{split}
\end{equation}
\end{theorem}
Theorem~\ref{theorem-2} indicates the infeasibility of optimizing the lower bound of error, which is determined by the dataset distribution, represented by A, X, and Y. 
The proof of Theorem~\ref{theorem-1} and Theorem~\ref{theorem-2} are provided in Appendix~\ref{proof}.

% \subsection{Challenges in optimizing \textbf{term (i)} and \textbf{term (ii)}}
% \label{app:challenge}
To optimize the model on the target graph, we encounter the following issues:
(C1) \textbf{Independent training of the GNNs.} A usual assumption is that the testing graph is the same as the training graph. This premise
requires independently training the structure learning model from
scratch for each graph dataset which leads to prohibitive computation costs and potential risks for serious over-fitting.
(C2) \textbf{Distribution alignment.} Obtaining relatively domain-invariant semantic information in a disentangled manner is challenging when dealing with source domains with significant distribution differences. In other words, aligning $\mathbb{P}^{e_{i}}_{S|Y}$ and $\mathbb{P}^{e_{j}}_{S|Y}$ in the second part of Eq.~\eqref{eq:upper_acc} becomes difficult.

\textbf{Addressing (C1) by training the structure learner.}
If we can learn from the $A$ and $X$ of source graphs how to capture structural information on the graph, we can then apply this ability to the target domain. Specifically, we decompose the representation mapping $g: \mathcal{A}, \mathcal{X} \to \mathbb{R}^s$ in Theorem~\ref{theorem-1} into three parts: $g=f_t\circ\text{GNN}\circ E_s$, where $f_t$ is a structure learner to refine the given structure (i.e., $A'=f_t(A,X)$), GNN embeds ($A',X$) and ($A,X$) to $R$, and $E_s$ learns the semantic factor, \textit{i.e.,} $S=E_s(R)$. Therefore, we can use this pre-trained $f_t$ and $E_S$ on the target graph instead of retraining the entire model.
\textbf{Addressing (C2) by feature disentanglement.} To ensure the final classification, we aim to decouple domain-invariant information, which has a consistent distribution across each domain. In a perfect decoupling scenario, the second part of Eq.~\eqref{eq:upper_acc} would approach zero.

\section{Experiments}
We apply \sysname{} to real-world datasets to investigate the effectiveness of domain generalization on graphs, which focuses on the following research questions.
\begin{itemize}[leftmargin=*]
    \item \textbf{RQ1:} Dose \sysname{} surpass the the state-of-the-art methods in the field of domain generalization on graphs?
    \item \textbf{RQ2:} To what extent does each critical component contribute to the overall performance of the \sysname{}?
    % \item \textbf{RQ3:} How sensitive is the model to the settings of its hyperparameters?
    \item \textbf{RQ3:} How does the representation learner improve the generalization of GNN?
\end{itemize}

\subsection{Experiment Setup}
\subsubsection{Datasets}
We utilize three real network datasets that
come with ground truth to verify the effectiveness of \sysname{}. 
Experiments are conducted on the multiple graphs contained within each dataset.
The statistical characteristics of these
networks are shown in Table~\ref{tab.data} and due to space constraints, we present the details of the dataset in the Appendix~\ref{dataset}.
\begin{table}[htbp]
%\footnotesize
% \tiny
 \setlength\tabcolsep{1.5pt}
  \caption{Key Characteristics of Datasets }
  % \vspace{-3mm}
  \label{tab.data}
  \centering
  \begin{tabular}{lccccc}
    \toprule
    \textbf{Dataset} &  \textbf{\# Nodes} & \textbf{\# Edges} & \textbf{\# Classes} & \textbf{\# Features} & \textbf{\# Domains}\\
    \midrule
    \textsc{Twitch-Explicit}~\cite{rozemberczki2021multi} &  20945 & 153,138 & 2 & 3170 & 7 (DE, ENGB, ES, FR, PTBR, RU, and TW)\\
    \textsc{Facebook-100}~\cite{traud2012social} & 131,924 & 1,590,655  & 2  & 12412  & 5 (Amh, Johns, Reed, Cornel, and Yale)\\
    \textsc{WebKB}~\cite{pei2020geom} & 617 & 1138 & 5 &  1703 &3 (Cornell, Texas, and Wis) \\
    \bottomrule
  \end{tabular}
  % \vspace{-3mm}
\end{table}

\subsubsection{Baselines}
We compare \sysname{} with graph domain generation methods (EERM~\cite{wu2022handling}, SRGNN~\cite{zhu2021shift}, FLOOD~\cite{liu2023flood}), data augmentation methods for graphs (Mixup~\cite{wang2021mixup}), meta-learning methods for graphs (GMeta~\cite{huang2020graph}, GraphGlow\cite{zhao2023graphglow}, MD-Gram~\cite{lin2023multi}) and a base method ERM.
Readers can refer to Appendix~\ref{baseline} for more details of baseline methods.

\subsubsection{Implementation Details}
\label{Implementation}
We establish $3$ different scenarios determined by whether the source and target graphs are derived from the same dataset.
\begin{itemize}[leftmargin=*]
    \item \textbf{S1T1}. Both source and target graphs originate from the same dataset. We sequentially test each graph for each dataset while training on the remaining ones.
    \item \textbf{S1T2}. The source graphs and target graphs are from different datasets. In particular, all graphs in the source are from the same dataset. For instance, we use $5$ graphs from \textsc{Facebook-100} for training and testing on the graphs of \textsc{Twitch-Explicit} and \textsc{WebKB}, separately. This approach can be similarly applied to other datasets.
    \item \textbf{S12T3}. The source graphs and target graphs are from different datasets. In particular, the source graphs for training are selected from different datasets. Here, we choose eight graphs from two distinct datasets for training (\textit{e.g.,} \textsc{Facebook-100} and \textsc{Twitch-Explicit}), and testing on the other dataset (\textsc{WebKB}).
\end{itemize}
We use GCN architectures as the GNN model in \sysname{}.
For all baseline models, we implemented them using the authors' provided source code and also set GCN as the backbone.
To reduce the time and space cost of the structure learner,  following the simplification of \cite{zhao2023graphglow}, we convert the sampling of 
 $A'$ in the structure learner to the product of the 
(NP)-dimensional matrix and its transpose. 
Due to the baseline methods' inability to adapt to varying feature and label dimensions, we employ zero-padding for feature dimensions and label expansion to standardize them after comparing different padding methods.
We report the experimental results for all scenarios, with the final result for each dataset derived from the mean of all graphs within that dataset. 
The results represent the average values obtained from $10$ runs for all the methods compared.

% \begin{table*}[htbp]
% \footnotesize
%  \setlength\tabcolsep{1.5pt}
%   \caption{experiment setting}
%   \label{tab.setting}
%   \centering
%   \begin{tabular}{l|c|c|c}
%     \toprule
%    \textbf{setting} & \textbf{training} &  \textbf{testing} & Table\\
%     \midrule
%     cross datasets & wb-Texas+wb-Cornell+wb-Wisconsin+fb-reed+fb-amherst+fb-jh+fb-cornell-fb-yale (WebKB+Fb-100) & twitch & Table~\ref{tab.Twitch_fb+web}\\
%     cross datasets& tw-PTBR+tw-TW+tw-RU+tw-ES+fb-amherst+fb-jh+fb-reed+fb-Cornell (Twitch+Fb-100) & webKB & Table~\ref{tab.webkb_Tw+fb}\\
%     cross datasets & tw-PTBR+tw-TW+tw-RU+tw-ES+tw-FR+wb-Cornell+web-Wisconsin+wb-Texas (WebKB+Twitch) & Fb-100 & Table~\ref{tab.fb_web+fb100}\\
%     cross dataset & Fb-100 & Twitch + webKB & Table~\ref{tab.twitch+web_fb}\\
%     cross dataset & Twitch & Fb-100 + webKB & Table~\ref{tab.fb+web_twitch} \\
%     cross dataset & webKB & Twitch + Fb-100 & Table~\ref{tab.twitch+fb_web}\\
%     In dataset & Fb-100  & Fb-100 & Table~\ref{tab.in-dataset}\\
%     In dataset & Twitch  & Twitch & Table~\ref{tab.in-dataset}\\
%     In dataset & WebKB & WebKB & Table~\ref{tab.in-dataset}\\
%     \bottomrule
%   \end{tabular}
% \end{table*}

\subsubsection{Hyper-parameter Settings}
The parameters of the structure learner include $\lambda$ (the weight on original graphs), $\alpha$ (the weight on smoothness constraints), and $\beta$ (the weight on sparsity constraints).
We adjust all values to fall within the range of $[0,1]$.
In meta-training, we set the update step as $5$.
In meta-testing, we set the update step in $\{1, 5, 10, 20, 30, 40\}$.
The learning rates for the inner and outer loops are set to $l_{in} = 1e^{-3}$ and $l_{out} = 1e^{-1}$, respectively.
\begin{table*}[t]
% \scriptsize
\tiny
 \setlength\tabcolsep{2pt}
  \caption{Test accuracy (\%) on \textsc{Twitch}, \textsc{FB-100} and \textsc{WebKB} where source and target graphs from same dataset. }
  % \vspace{-3mm}
  \label{tab.in-dataset}
  \centering
  \begin{tabular}{l|ccccccc|c|ccccc|c}
   \hline
   \multirow{2}{*}{\textbf{Methods}} & \multicolumn{8}{c|}{\textbf{S1T1 (\textsc{Twitch} $\rightarrow$ \textsc{Twitch})}} & \multicolumn{6}{c}{\textbf{S1T1 (\textsc{FB-100} $\rightarrow$ \textsc{FB-100})}}  \\
   \cline{2-15}
   ~ & \textbf{PTBR}  & \textbf{TW} & \textbf{RU} & \textbf{ES} & \textbf{FR} & \textbf{ENGB} & \textbf{DE} & \textbf{Avg} & \textbf{Amherst}  & \textbf{Johns} & \textbf{Reed} & \textbf{Cornell} & \textbf{Yale} & \textbf{Avg} \\
    \hline
    GraphGlow~\cite{zhao2023graphglow} & $65.4 \pm 0.6$ & $60.7 \pm 0.2$ & $75.4 \pm 0.5$ & $70.7 \pm 0.5$ & $63.1 \pm 0.2$ &$54.5 \pm 0.1$ & $60.4 \pm 0.9$ & $64.3$ & $53.0 \pm 0.5$ & $50.1 \pm 0.9$ & $61.2 \pm 1.1$  & $51.6 \pm 0.3$ & $50.9 \pm 0.8$ & $53.4 $ \\
    MD-Gram~\cite{lin2023multi} & $61.3 \pm 1.2$ &$59.2 \pm 1.3$ &$70.1 \pm 0.9$ & $65.7 \pm 0.4$ & $60.9 \pm 0.6$ &$53.3 \pm 0.8$ &$56.9 \pm 1.0$ & $61.1$ & $51.9 \pm 0.3$ & $50.1 \pm 0.8$ &$59.9 \pm 0.5$ &$51.1 \pm 0.4$ & $49.0 \pm 0.7$ & $52.4$ \\
    GMeta~\cite{huang2020graph} &$60.0 \pm 0.1$ & $59.4 \pm 0.6$ & $61.2 \pm 0.4$ & $63.5 \pm 0.2$ & $60.2 \pm 0.4$ & $49.8 \pm 0.7$ & $53.6 \pm 0.1$ & $58.2$ &$46.4 \pm 0.8$ & $44.5 \pm 1.0$ & $48.9 \pm 0.5$ & $40.2 \pm 0.1$ & $46.8 \pm 0.9$ & $45.4$  \\
    FLOOD~\cite{liu2023flood} & $57.0 \pm 0.3$ & $49.0 \pm 0.4$ &$46.1 \pm 0.2$ & $55.3 \pm 0.4$ & $ 50.4 \pm 0.2$ &$51.3 \pm 0.3$ & $53.1 \pm 0.4$ & $51.7$ & $49.8\pm 0.3$ & $47.1 \pm 0.4$ & $48.9 \pm 0.6$ & $41.0 \pm 0.3$ & $45.1 \pm 0.2$ & $46.4$  \\
    EERM~\cite{wu2022handling} & $65.3 \pm 0.0$ & 60.7 $\pm$ 0.0 & $72.7 \pm 3.8$ & $70.9 \pm 0.2$ & $59.3 \pm 4.4$ & $46.6 \pm 0.2$ & $51.0\pm 1,1$ & $60.9 $ & $53.9 \pm 0.7$ & $50.4 \pm 1.7$ & $51.8 \pm 0.9$  &$52.6 \pm 1.1$ & $52.9 \pm 1.3$ & $52.3$  \\
    SRGNN~\cite{zhu2021shift} & $38.8 \pm 2.6$ & $44.3 \pm 3.7$ & $58.8\pm 1.1$ & $53.8 \pm 1.5$ & $48.4\pm 0.4$ & $40.3\pm 2.0$ & $44.9 \pm 0.1$ & $47.0 $ & $46.1 \pm 1.3$ & $47.9 \pm 0.6$ & $49.5\pm 1.0$ & $46.1\pm 1.4$ &$48.8 \pm 0.9$ & $47.7$ \\
    Mixup~\cite{zhang2017mixup} & $45.1\pm 1.2$ & $45.3 \pm 0.9$ & $39.9 \pm 1.5$ & $48.1 \pm 1.3$ & $42.6\pm 1.3$ & $40.9\pm 0.6$ & $46.2 \pm 0.8$ & $44.0 $ & $45.8\pm 0.6$ & $46.1 \pm 0.7$ & $48.4 \pm 0.2$ & $47.2 \pm 0.4$ & $47.1 \pm 0.2$ & $46.9$  \\
    ERM~\cite{kipf2016semi} & $65.0 \pm 0.2$ & $58.6 \pm 2.9$ & $66.8 \pm 3.5$ &$70.1 \pm 0.1$ & $63.0 \pm 0.1$ & $45.2 \pm 0.0$ & $42.1 \pm 0.9$ & $58.7$ & $44.6 \pm 0.5$ & $46.9 \pm 0.9$ & $47.8 \pm 1.2$ &$44.3 \pm 1.0$  & $50.1 \pm 0.7$ & $46.7$ \\
    \hline
    \sysname{} & \underline{$67.8 \pm 0.3$} & \underline{$61.5 \pm 0.2$} & \underline{$76.0 \pm 0.1$} & \underline{$71.4 \pm 0.1$} & \underline{$63.7 \pm 0.2$} & \underline{$55.0 \pm 0.2$} & \underline{$60.5 \pm 0.1 $} &\underline{$65.1$ } & \underline{$55.2\pm 0.1$} & \underline{$51.3 \pm 0.1$ }& \underline{$62.9 \pm 0.2$} & \underline{$53.8 \pm 0.2$} & \underline{$54.0 \pm 0.2$} & \underline{$55.5$} \\
    \sysname{}-ind & \textbf{67.9} $\pm$ \textbf{0.8} & \textbf{62.0} $\pm$ \textbf{0.1}  & \textbf{76.1} $\pm$ \textbf{0.1} & \textbf{71.5} $\pm$ \textbf{0.3} & \textbf{64.8} $\pm$ \textbf{0.2} & \textbf{55.5} $\pm$ \textbf{0.3}  & \textbf{61.9} $\pm$ \textbf{0.5}&  \textbf{65.7} & \textbf{55.5} $\pm$ \textbf{0.1} & \textbf{51.4} $\pm$ \textbf{0.3}  & \textbf{64.2} $\pm$ \textbf{0.1} & \textbf{54.2} $\pm$ \textbf{0.4} & \textbf{54.3} $\pm$ \textbf{0.5}  &\textbf{55.9}  \\
    \hline
  \end{tabular}
  \vspace{-4mm}
\end{table*}
\begin{table*}[t]
% \scriptsize
\tiny
 \setlength\tabcolsep{2pt}
  % \vspace{-3mm}
  \centering
  \begin{tabular}{l|ccc|c}
   \hline
   \multirow{2}{*}{\textbf{Methods}}  & \multicolumn{4}{c}{\textbf{S1T1 (\textsc{WebKB} $\rightarrow$ \textsc{WebKB})}} \\
   \cline{2-5}
   ~ & \textbf{Texas}  & \textbf{Cornell} & \textbf{Wis} & \textbf{Avg} \\
    \hline
    GraphGlow~\cite{zhao2023graphglow} & \underline{$57.3 \pm 2.1$} & $44.8 \pm 1.5$ & $46.3 \pm 0.6$  & $49.5$ \\
    MD-Gram~\cite{lin2023multi} & $55.8 \pm 0.6$ &$43.9 \pm 0.7$ &$45.2 \pm 0.7$ & $48.3$\\
    GMeta~\cite{huang2020graph} & $42.3 \pm 0.7$ & $20.5 \pm 1.9$ & $40.1 \pm 0.4$ & $34.3$ \\
    FLOOD~\cite{liu2023flood} & $22.5 \pm 0.6$ & $18.7 \pm 0.2$ & $15.6 \pm 0.1$ & $18.9$ \\
    EERM~\cite{wu2022handling} & $31.1 \pm 2.1$ & $19.2 \pm 1.3$ & $7.7 \pm 1.7$  & $19.3$ \\
    SRGNN~\cite{zhu2021shift} & $14.3 \pm 2.8$ & $15.8 \pm 0.6$ & $18.7\pm 2.5$ & $16.3$\\
    Mixup~\cite{zhang2017mixup} & $14.0\pm 1.2$ & $16.2 \pm 1.0$ & $17.9 \pm 1.6$ & $16.0 $ \\
    ERM~\cite{kipf2016semi} & $42.8 \pm 3.9$ & $12.8 \pm 2.0$ & $13.5 \pm 1.2$ &$23.0$\\
    \hline
    \sysname{} & $56.6 \pm 0.2$ & \underline{$47.3 \pm 0.3$} & \underline{$51.7 \pm 0.3$} & \underline{$51.9$}\\
    \sysname{}-ind & \textbf{59.7} $\pm$ \textbf{0.3} & \textbf{49.6} $\pm$ \textbf{0.2}  & \textbf{54.0} $\pm$ \textbf{0.2} & \textbf{54.4} \\
    \hline
  \end{tabular}
  % \vspace{-3mm}
\end{table*}

\begin{table}[t]
% \begin{minipage}{0.695\textwidth}
% \begin{table*}[!t]
\tiny
\centering
  \setlength\tabcolsep{2pt}
  \caption{Test accuracy (\%) on \textsc{FB-100} and \textsc{WebKB} where all source graphs from \textsc{Twitch}. }
    % \vspace{-3mm}
  \label{tab.fb+web_twitch}
  \centering
  \begin{tabular}{l|ccccc|c|ccc|c}
    \hline
   \multirow{2}{*}{\textbf{ Methods}} & \multicolumn{6}{c|}{\textbf{S1T2 ( Twitch $\rightarrow$ FB-100 )}} & \multicolumn{4}{c}{\textbf{S1T2 ( Twitch $\rightarrow$ WebKB )}} \\
   \cline{2-11}
   ~& \textbf{Amherst}  & \textbf{Johns} & \textbf{Reed} & \textbf{Cornell} & \textbf{Yale} & \textbf{Avg} & \textbf{Texas}  & \textbf{Cornell} & \textbf{Wis} & \textbf{Avg}\\
    \hline
    GraphGlow~\cite{zhao2023graphglow} & $52.9 \pm 1.5$ & $52.3 \pm 1.2 $ & $60.7 \pm 1.7$  & $51.2 \pm 1.0$ & $43.5 \pm 1.2$ & $52.1$ & $53.0 \pm 1.1$ & $44.8 \pm 1.3$ & $47.0 \pm 0.9$  & $48.3$ \\
    MD-Gram~\cite{lin2023multi} & $50.2 \pm 0.3$ & $47.9 \pm 0.4$ &$62.8 \pm 0.4$ &$49.0 \pm 0.2$ &$42.7 \pm 0.7$ &$50.5$ &$53.8 \pm 0.2$ &$41.6 \pm 0.3$ &$44.2 \pm 0.3$ & $46.5$\\
    GMeta~\cite{huang2020graph} & $20.1 \pm 0.8$ & $14.5 \pm 1.7$ & $16.0 \pm 1.6 $ & $16.9 \pm 1.4$ & $16.3 \pm 2.1$ & $16.8 $ &$19.8 \pm 1.7$ & $16.0 \pm 1.4$ & $16.2 \pm 1.3 $ & $17.3$\\
    FLOOD~\cite{liu2023flood} & $11.1 \pm 0.3$ & $14.0 \pm 0.4$ & $11.9 \pm 0.3$ &$14.2 \pm 0.2$ & $14.3 \pm 0.5$ & $13.1$ & $19.3 \pm 0.2$ &$17.2 \pm 0.4$ &$11.0 \pm 0.3$ & $15.8$\\
    EERM~\cite{wu2022handling} & $10.9 \pm 2.4$ & $14.3 \pm 3.1$ & $12.8 \pm 2.0$ & $13.0 \pm 2.6$ & $14.4 \pm 3.0$ & $13.1 $ & $20.8 \pm 2.3$ & $18.0 \pm 2.5$ & $11.2 \pm 1.9$ & $16.7$\\
    SRGNN~\cite{zhu2021shift} & $11.7 \pm 1.3$ & $15.0 \pm 1.8$ & $11.6 \pm 1.5$ & $12.7 \pm 1.4$ & $13.9 \pm 2.6$  & $13.0 $ & $16.7 \pm 1.9$ & $11.4 \pm 0.6$ & $10.8 \pm 1.5$ & $13.0 $\\
    Mixup~\cite{zhang2017mixup} &$10.8 \pm 1.7$ &$12.9 \pm 1.0$ &$12.1 \pm 1.3$ &$11.3 \pm 1.2$ &$13.2 \pm 1.5$ &$12.1$ & $17.5 \pm 1.1$ & $10.9\pm 1.4$ & $11.1 \pm 1.3$ & $13.2 $\\
    ERM~\cite{kipf2016semi} & $ 18.9 \pm 2.0$ & $10.2 \pm 1.6$ & $11.6 \pm 2.2$ & $19.0 \pm 2.1$ & $11.9 \pm 1.8 $ & $ 12.3 $ & $20.2 \pm 1.4$ & $15.5 \pm 2.5$ & $10.6 \pm 4.4$ & $15.4 $\\
    \hline
     \sysname{} &\underline{$55.5 \pm 0.4$} & \underline{$51.4 \pm 0.5$} & \underline{$64.1 \pm 0.2$ }& \underline{$52.4 \pm 0.2$} & \underline{$53.5 \pm 0.3$} & \underline{$55.4 $ } & \underline{$58.1\pm 0.1$} & \underline{$50.4 \pm 0.2$} & \underline{$51.1 \pm 0.2$ }& \underline{$53.2$} \\
    \sysname{}-ind & \textbf{55.3} $\pm$ \textbf{0.5} & \textbf{51.9} $\pm$ \textbf{1.2}  & \textbf{64.0} $\pm$ \textbf{1.1} & \textbf{53.5} $\pm$ \textbf{0.7} & \textbf{54.0} $\pm$ \textbf{0.8}  &\textbf{55.7} & \textbf{59.5} $\pm$ \textbf{0.2} & \textbf{48.9} $\pm$ \textbf{0.1}  & \textbf{51.5} $\pm$ \textbf{0.1} & \textbf{53.3} \\
    \hline
  \end{tabular}
% \vspace{-2mm}
% \end{table*}
\end{table}
% \hfill

\subsection{Results}

\begin{table}[t]
% \begin{minipage}{0.695\textwidth}
% \begin{table*}[!t]
\tiny
  \setlength\tabcolsep{2pt}
  \caption{Test accuracy (\%) on \textsc{Twitch} and \textsc{WebKB} where source graphs all from \textsc{FB-100}.}
   % \vspace{-3mm}
  \label{tab.twitch+web_fb}
  \centering
  \begin{tabular}{l|ccccccc|c|ccc|c}
   \hline
   \multirow{2}{*}{\textbf{Methods}} & \multicolumn{8}{c|}{\textbf{S1T2 ( FB-100 $\rightarrow$ Twitch )}} & \multicolumn{4}{c}{\textbf{S1T2 ( FB-100 $\rightarrow$ WebKB )}}\\
   \cline{2-13}
   ~ &\textbf{PTBR}  & \textbf{TW} & \textbf{RU} & \textbf{ES} & \textbf{FR} & \textbf{ENGB} & \textbf{DE} & \textbf{Avg} &  \textbf{Texas}  & \textbf{Cornell} & \textbf{Wis} & \textbf{Avg} \\
    \hline
 GraphGlow~\cite{zhao2023graphglow} & $65.4 \pm 0.4$ & $ 60.7\pm 0.6$ & $ 75.4\pm 0.9$ & 70.7$ \pm 0.9$ & $63.1 \pm 0.7$ & $54.5 \pm 0.6$ & $60.4 \pm 0.8$ & $64.3 $ & \underline{$59.0\pm 0.5$} & $ 44.8\pm 0.4$ & $46.6 \pm 0.9$  & $ 50.1$ \\
  MD-Gram~\cite{lin2023multi} & $64.8 \pm 0.3$ & $58.7 \pm 0.4$ &$73.0 \pm 0.1$ &$71.1 \pm 0.9 $ &$62.0 \pm 0.4$ &$52.9 \pm 0.5$ &$60.1 \pm 0.5$ &$63.2$ &$54.1 \pm 0.4$ &$44.8 \pm 0.7$ &$45.0 \pm 0.2$ &$48.0$\\
    GMeta~\cite{huang2020graph}  &$10.5 \pm 1.9$ & $26.6 \pm 1.4$ & $14.8 \pm 2.2$ & $7.1 \pm 1.4$ & $23.7 \pm 2.4$ & $13.3 \pm 1.0$ & $15.8 \pm 2.5$ &$ 15.9 $ &$20.0 \pm 1.9$ & $15.8 \pm 2.0$ & $18.6 \pm 1.7$ & $18.1 $ \\
    FLOOD~\cite{liu2023flood} & $11.4 \pm 0.3$ & $24.3 \pm 0.2$ & $18.1 \pm 0.2$  & $10.2 \pm 0.6$ & $24.8 \pm 0.1$ & $11.9 \pm 0.4$ &$16.0 \pm 0.4$ & $19.1$ & $18.3  \pm 0.1$ &$16.9 \pm 0.3$ & $11.0 \pm 0.4$ & $15.4$\\
    EERM~\cite{wu2022handling}  &$11.4 \pm 2.4$ & $29.8 \pm 2.5$ & $17.0 \pm 1.8$ & $6.3 \pm 2.7$ & $25.0 \pm 2.9$ & $12.3 \pm 2.4$ & $16.9 \pm 2.9$ &$ 17.0 $ & $17.5 \pm 2.8$ & $12.6 \pm 2.9$ & $10.8 \pm 1.3$ & $13.6 $ \\
    SRGNN~\cite{zhu2021shift}  &$18.3 \pm 1.6$ & $15.9 \pm 2.0$ & $15.0 \pm 1.8$ & $18.1 \pm 1.2$ & $14.7 \pm 2.0$ & $10.5 \pm 1.8$ & $11.4 \pm 1.4$ &$15.0  $ & $18.0 \pm 1.1$ & $10.7 \pm 1.4$ & $11.5 \pm 1.2$ & $13.4 $\\
    Mixup~\cite{zhang2017mixup} &$19.7 \pm 1.0$ & $27.5 \pm 1.9$ & $16.4 \pm 1.2$ & $17.0 \pm 1.8$ & $19.6 \pm 1.6$ & $12.0 \pm 2.1$ & $13.8 \pm 1.7$ &$ 18.1$ & $19.7 \pm 1.5$ & $12.6 \pm 1.0$ & $12.8 \pm 0.9$ & $15.0 $\\
    ERM~\cite{kipf2016semi}  & $16.7 \pm 1.3$ & $27.0 \pm 2.9$ &$15.0 \pm 3.4$ &$13.5 \pm 2.6$ &$14.3 \pm 2.2$ & $15.7 \pm 3.1$ & $14.3 \pm 2.0$ & $15.9 $& $33.5 \pm 3.7$ &$38.2 \pm 3.0$ & $30.1 \pm 1.5$ &$33.9 $\\
   \hline
     \sysname{} & \underline{$66.7 \pm 0.1$} & \underline{$61.2\pm 0.2$} & \underline{$76.5\pm 0.2$} & \underline{$71.9 \pm 0.2$} & \underline{$64.0 \pm 0.3$} & \underline{$55.0 \pm 0.1$} & \underline{$60.8 \pm 0.3$ } & \underline{$65.2 $} &$56.6 \pm 0.2$ & \underline{$48.1 \pm 0.3$} & \underline{$48.9 \pm 0.2$ } & \underline{$51.2 $}\\
    \sysname{}-ind & \textbf{67.1} $\pm$ \textbf{0.2} & \textbf{61.8} $\pm$ \textbf{0.1}  & \textbf{76.5} $\pm$ \textbf{0.1} & \textbf{71.9} $\pm$ \textbf{0.3} & \textbf{64.0} $\pm$ \textbf{0.2} & \textbf{55.6} $\pm$ \textbf{0.1}  & \textbf{61.1} $\pm$ \textbf{0.2}&  \textbf{65.4} & \textbf{60.5} $\pm$ \textbf{0.4} & \textbf{48.3} $\pm$ \textbf{0.2}  & \textbf{52.3 } $\pm$ \textbf{0.2} & \textbf{53.7} \\
    \hline
  \end{tabular}
    % \vspace{-1mm}
% \end{table*}
\end{table}
% \hfill

\begin{table}[t]
% \scriptsize
\tiny
  \setlength\tabcolsep{2pt}
  \caption{Test accuracy (\%) on \textsc{Twitch} and \textsc{FB-100} where source graphs all from \textsc{WeBKB}.}
   % \vspace{-3mm}
  \label{tab.twitch+fb_web}
  \centering
  \begin{tabular}{l|ccccccc|c|ccccc|c}
  \hline
  \multirow{2}{*}{\textbf{Methods}} & \multicolumn{8}{{c|}}{\textbf{S1T2 (WebKB $\rightarrow$ Twitch)}} & \multicolumn{6}{{c}}{\textbf{S1T2 (WebKB $\rightarrow$ FB-100)}} \\
  \cline{2-15}
   ~ & \textbf{PTBR}  & \textbf{TW} & \textbf{RU} & \textbf{ES} & \textbf{FR} & \textbf{ENGB} & \textbf{DE} & \textbf{Avg} & \textbf{Amherst}  & \textbf{Johns} & \textbf{Reed} & \textbf{Cornell} & \textbf{Yale} & \textbf{Avg}\\
    \hline
    GraphGlow~\cite{zhao2023graphglow} & $65.4 \pm 1.1$ & $ 60.7\pm 1.3$ & $ 75.4\pm 0.9$ & 70.7$ \pm 0.3$ & $63.1 \pm 1.2$ & $54.5 \pm 0.2$ & \underline{$60.4 \pm 0.5$} & $64.3$ & $52.7 \pm 1.8$ & $51.3 \pm 1.0 $ & $61.7 \pm 1.2$  & $51.3 \pm 1.5$ & $52.4 \pm 1.3$ & $53.8$ \\
    MD-Gram~\cite{lin2023multi} &$64.3 \pm 0.4$ &$57.8 \pm 0.3$ &$72.0 \pm 0.4$ &$70.7 \pm 0.8$ &$61.4 \pm 0.5$ &$52.2 \pm 0.6$ &$58.9 \pm 0.5$ &$62.5$ & $51.1 \pm 0.4$ & $48.0 \pm 0.3$ &$62.9 \pm 0.6$ & $48.7 \pm 0.3$ &$44.2 \pm 0.6$ &$51.0$ \\
    GMeta~\cite{huang2020graph}  & $12.7 \pm 2.1$ & $23.5 \pm 1.3$ & $28.9 \pm 1.4$ & $30.8 \pm 1.7$ & $13.0 \pm 2.1$ &$15.2 \pm 1.8$ & $19.3\pm 1.2$ &$22.3$ & $29.7 \pm 1.5$ & $26.6 \pm 2.0$ & $19.8 \pm 0.9 $ & $21.5 \pm 1.8$ & $20.4 \pm 1.2$ & $23.6 $\\
    FLOOD~\cite{liu2023flood} & $ 10.0 \pm 0.3$ & $28.5 \pm 0.4$ & $28.1 \pm 0.3$ & $31.4 \pm 0.2$ & $20.9 \pm 0.1$ &$ 11.8 \pm 0.2$ & $16.7 \pm 0.5$ & $21.0$ & $40.7 \pm 1.0$ &$39.9 \pm 0.4$ &$39.0 \pm 0.3$ & $45.5 \pm 0.4$ &$41.2 \pm 0.8$ &$41.3$ \\
    EERM~\cite{wu2022handling} & $10.1 \pm 3.3$ &$31.1 \pm 2.5$ &$30.8 \pm 2.7$ &$34.1\pm 2.9$ & $22.8 \pm 3.5$ &$13.9 \pm 2.1$ & $15.4 \pm 0.8$ &$22.6 $ &$42.8 \pm 2.7$ & $41.4 \pm 2.6$ & $40.6 \pm 2.8$ & $49.1 \pm 3.5$ &$43.8 \pm 2.6$ &$43.5$ \\
    SRGNN~\cite{zhu2021shift}  & $11.2 \pm 0.9$ &$29.6 \pm 1.4$ & $28.5 \pm 1.7$ &$31.9\pm 2.0$ & $24.1 \pm 2.2$ & $13.4 \pm 1.3$ & $16.1 \pm 1.0$ &$22.1$ & $21.2 \pm 1.8$ & $19.6 \pm 1.7$ & $18.3 \pm 0.9$ & $20.9 \pm 1.7$ & $21.6 \pm 1.5$  & $20.3$ \\
    Mixup~\cite{zhang2017mixup} & $12.0 \pm 1.3$ &$27.5 \pm 1.7$ & $27.9 \pm 1.6$ & $31.4\pm 1.5$ & $25.0 \pm 1.9$ & $14.2 \pm 1.6$ & $16.3 \pm 1.0$ & $22.0 $ &$32.6 \pm 2.0$ &$31.7 \pm 1.2$ & $29.9 \pm 0.8$ & $32.0 \pm 1.9$ & $30.1 \pm 1.6$ &$31.3 $ \\
    ERM~\cite{kipf2016semi}  &$34.5 \pm 2.9$ & $10.5 \pm 0.4$ &$41.1 \pm 4.8$  & $20.7 \pm 3.3$ & $16.8 \pm 0.2$ & $12.6 \pm 0.2$ & $15.0\pm 0.1$ &$21.7$ &$40.9 \pm 2.6$ & $46.0 \pm 2.1$ &$40.5 \pm 2.4$ & $42.1 \pm 2.6$ & $44.4 \pm 3.8$ & $42.7$ \\
   \hline
    
    \sysname{}& \underline{$66.2 \pm 0.3$ 
} & \underline{$61.8 \pm 0.1$} & \underline{$75.9 \pm 0.2$} & \underline{$71.3 \pm 0.2$} & \underline{$63.9 \pm 0.1$} & \underline{$55.1 \pm 0.3$} &$60.3 \pm 0.3$ & \underline{$64.9$} & \underline{$55.7 \pm 0.1$ }& \underline{$51.5 \pm 0.2$} & \underline{$65.0 \pm 0.1$ }& \underline{$53.1 \pm 0.4$} & \underline{$53.5 \pm 0.2$} &\underline{$55.8$}\\
  \sysname{}-ind & \textbf{66.8} $\pm$ \textbf{0.5} & \textbf{61.8} $\pm$ \textbf{0.2}  & \textbf{76.5} $\pm$ \textbf{0.2} & \textbf{71.8} $\pm$ \textbf{0.1} & \textbf{63.9} $\pm$ \textbf{0.1} & \textbf{55.2} $\pm$ \textbf{0.4}  & \textbf{61.1} $\pm$ \textbf{0.2}&  \textbf{65.3} & \textbf{55.7} $\pm$ \textbf{0.3} & \textbf{51.7} $\pm$ \textbf{0.6}  & \textbf{65.0} $\pm$ \textbf{0.4} & \textbf{53.9} $\pm$ \textbf{0.6} & \textbf{53.6} $\pm$ \textbf{0.5}  &\textbf{56.0} \\
    \hline
  \end{tabular}
% \vspace{-3mm}
\end{table}

\begin{table}[t]
% \scriptsize
\tiny
  \setlength\tabcolsep{2pt}
  \caption{Test accuracy (\%) on \textsc{Twitch}, \textsc{FB-100} and \textsc{WebKB} where source and target graphs from different datasets.}
  % \vspace{-3mm}
  \label{tab.cross-train-2}
  \centering
   \begin{tabular}{l|ccccccc|c|ccccc|c}
   \hline
   \multirow{2}{*}{\textbf{Methods}} & \multicolumn{8}{c|}{\textbf{S12T3 (FB-100 + WebKB  $\rightarrow$ Twitch)}} & \multicolumn{6}{c}{\textbf{S12T3 (Twitch + WebKB  $\rightarrow$ FB-100)}} \\
   \cline{2-15}
   ~ & \textbf{PTBR}  & \textbf{TW} & \textbf{RU} & \textbf{ES} & \textbf{FR} & \textbf{ENGB} & \textbf{DE} & \textbf{Avg} & \textbf{Amherst41}  & \textbf{Johns} & \textbf{Reed} & \textbf{Cornell} & \textbf{Yale} & \textbf{Avg}  \\
   \hline    GraphGlow~\cite{zhao2023graphglow} & $65.4 \pm 0.5$ & $ 60.7\pm 0.4$ & $ 75.4\pm 0.4$ & $70.7 \pm 0.4$ & $63.1 \pm 0.3 $ & $54.5 \pm 1.0$ & $60.4 \pm 0.7$ & $64.3 $ & $53.1 \pm 0.8$ & $47.4 \pm 1.2 $ & $63.2 \pm 1.1$  & $50.9 \pm 1.1$ & $43.4 \pm 1.1$ & $51.6 $ \\
    MD-Gram~\cite{lin2023multi} & $65.1 \pm 0.3$ & $60.9 \pm 0.1$ &$73.9 \pm 0.2$ & $71.0 \pm 0.2$ &$62.6 \pm 0.2$ &$53.8 \pm 0.1$ &$60.2 \pm 0.4$ & $63.9$ &$50.8 \pm 0.7$ &$48.3 \pm 0.4$ &$63.3 \pm 0.5$ &$49.9 \pm 0.4$ &$43.1 \pm 0.7$ & $51.1$ \\
    GMeta~\cite{huang2020graph} & $31.9 \pm 1.8$ & $24.0 \pm 1.2$ & $27.6 \pm 1.8$ & $31.0 \pm 2.1$ & $22.8 \pm 1.6$ & $26.0 \pm 1.1$ & $20.2 \pm 1.9 $ &$26.2$ &$21.3 \pm 2.1$ &$23.2 \pm 1.7$ & $19.9 \pm 1.6$ & $22.2 \pm 1.7$ &$21.5 \pm 1.4$ &$21.6 $ \\
    FLOOD~\cite{liu2023flood} & $24.9 \pm 0.6$ &$11.7 \pm 0.4$ & $24.1 \pm 0.7$ & $15.0 \pm 0.3$ & $13.9 \pm 0.4$ &$13.0 \pm 0.3$ &$14.1 \pm 0.4$ &$16.7$ & $19.1 \pm 0.1$ &$16.5 \pm 0.3$ &$12.0 \pm 0.3$ &$25.6 \pm 0.4$ &$15.8 \pm 0.2$ &$17.8$ \\
    EERM~\cite{wu2022handling}  & $26.5 \pm 2.1$ & $12.3 \pm 3.1$ &$25.8 \pm 2.8$ & $15.4 \pm 2.9$ & $ 14.8 \pm 2.8$& $13.5 \pm 3.0$ & $14.6 \pm 2.7$ & $17.6 $ & $19.9 \pm 4.1$ & $17.1 \pm 2.9$ & $12.1 \pm 4.3$ & $26.6 \pm 3.1$ & $16.9 \pm 2.2$ & $18.5 $ \\
    SRGNN~\cite{zhu2021shift}  & $11.3 \pm 1.2$ &$28.4 \pm 1.0$ & $27.7 \pm 2.1$ & $29.6 \pm 1.2$ & $25.3 \pm 1.6$ & $12.7 \pm 1.9$ & $17.3 \pm 1.5$ & $21.7 $ & $18.2 \pm 1.8$ & $18.0 \pm 1.6$ & $10.6 \pm 2.3$ & $24.9 \pm 1.8$ & $15.5 \pm 1.3$ & $17.4 $ \\
    Mixup~\cite{zhang2017mixup} & $12.3 \pm 1.7$ & $25.5 \pm 1.4$ & $26.9 \pm 2.0$ & $30.4\pm 1.2$ & $24.0 \pm 0.7$ & $13.1 \pm 0.5$ & $17.5 \pm 1.3$ & $21.4$ & $16.3 \pm 1.6$ & $13.6 \pm 1.9$ & $13.2 \pm 2.2$ & $22.5 \pm 1.4$ & $14.2 \pm 1.8$ & $16.0 $ \\
    ERM~\cite{kipf2016semi}  &$26.1 \pm 2.9$ & $15.0 \pm 2.6$ & $6.4 \pm 2.1$ &$13.8 \pm 0.4$ & $17.4 \pm 2.8$ &$15.2 \pm 3.6$ &$14.0 \pm 2.6$ &$15.6 $ &$19.3 \pm 0.1$ &$24.5 \pm 3.9$ &$10.5 \pm 2.7$ & $25.0 \pm 2.1$ & $12.7 \pm 2.8$ & $18.4 $ \\
   \hline
    \sysname{}&\underline{$66.6 \pm 0.4$} & \underline{$61.4 \pm 0.3$} & \underline{$75.8 \pm 0.3$} & \underline{$71.3 \pm 0.2$} & \underline{$63.2 \pm 0.3$} & \underline{$55.1 \pm 0.5$} & \underline{$62.3 \pm 0.5$} & \underline{$65.1$} &\underline{$55.5 \pm 0.3$} & \underline{$51.5 \pm 0.5$} & \underline{$64.5 \pm 0.5$} & \underline{$53.2 \pm 0.1$ }& \underline{$53.6 \pm 0.2$} & \underline{$55.7$} \\
     \sysname{}-ind  & \textbf{68.3} $\pm$ \textbf{0.6} & \textbf{62.9} $\pm$ \textbf{0.4}  & \textbf{77.1} $\pm$ \textbf{0.5} & \textbf{72.6} $\pm$ \textbf{0.5} & \textbf{65.0} $\pm$ \textbf{0.2} & \textbf{56.1} $\pm$ \textbf{0.6}  & \textbf{63.0} $\pm$ \textbf{0.5}&  \textbf{66.4} & \textbf{55.7} $\pm$ \textbf{0.5} & \textbf{52.0} $\pm$ \textbf{0.6}  & \textbf{64.6} $\pm$ \textbf{0.8} & \textbf{54.0} $\pm$ \textbf{0.2} & \textbf{54.2} $\pm$ \textbf{0.3}  &\textbf{56.1}  \\
    \hline
  \end{tabular}
  \vspace{-4mm}
\end{table}
\begin{table}[t]
% \scriptsize
\tiny
  \setlength\tabcolsep{2pt}
  % \vspace{-3mm}
  \centering
   \begin{tabular}{l|ccc|c}
   \hline
   \multirow{2}{*}{\textbf{Methods}} & \multicolumn{4}{c}{\textbf{S12T3 (FB-100 + Twitch  $\rightarrow$ WebKB)}} \\
   \cline{2-5}
   ~ & \textbf{Texas}  & \textbf{Cornell} & \textbf{Wis} & \textbf{Avg} \\
   \hline    GraphGlow~\cite{zhao2023graphglow} & $55.2 \pm 0.9$ & $44.8 \pm 1.2$ & $45.4 \pm 1.0$  & $48.5 $\\
    MD-Gram~\cite{lin2023multi} & $55.4 \pm 0.5$ &$45.2 \pm 0.3$ &$45.1 \pm 0.6$ &$48.6$\\
    GMeta~\cite{huang2020graph} & $23.2 \pm 1.4$ & $21.9 \pm 1.9$ & $19.6 \pm 2.0$ & $21.6 $ \\
    FLOOD~\cite{liu2023flood} & $19.7 \pm 0.3$ &$18.1 \pm 0.4$ &$15.5 \pm 0.4$ & $17.8$\\
    EERM~\cite{wu2022handling}  & $20.7 \pm 0.0$ & $18.3 \pm 0.4$ & $13.9 \pm 0.0$ & $17.6$\\
    SRGNN~\cite{zhu2021shift}  & $19.2 \pm 1.8$ &$15.7 \pm 1.6$ & $14.0 \pm 1.3$ &$16.3 $\\
    Mixup~\cite{zhang2017mixup} & $18.1 \pm 1.6$ & $13.9 \pm 2.1$ & $13.2 \pm 1.2$ & $15.1 $\\
    ERM~\cite{kipf2016semi}  & $21.3 \pm 0.9$ &$16.4 \pm 2.8$ &$15.2 \pm 2.0$ & $17.3 $\\
   \hline
    \sysname{}& \underline{$58.2 \pm 0.1$} & \underline{$50.4 \pm 0.3$} & \underline{$52.3 \pm 0.3$} & \underline{$53.6$}\\
     \sysname{}-ind  & \textbf{62.0} $\pm$ \textbf{0.2} & \textbf{50.6} $\pm$ \textbf{0.5}  & \textbf{51.2} $\pm$ \textbf{0.4} & \textbf{54.6} \\
    \hline
  \end{tabular}
  % \vspace{-3mm}
\end{table}

% \begin{minipage}{0.3\textwidth}
\begin{figure}[t]
    \centering
  \includegraphics[width=0.7\linewidth]{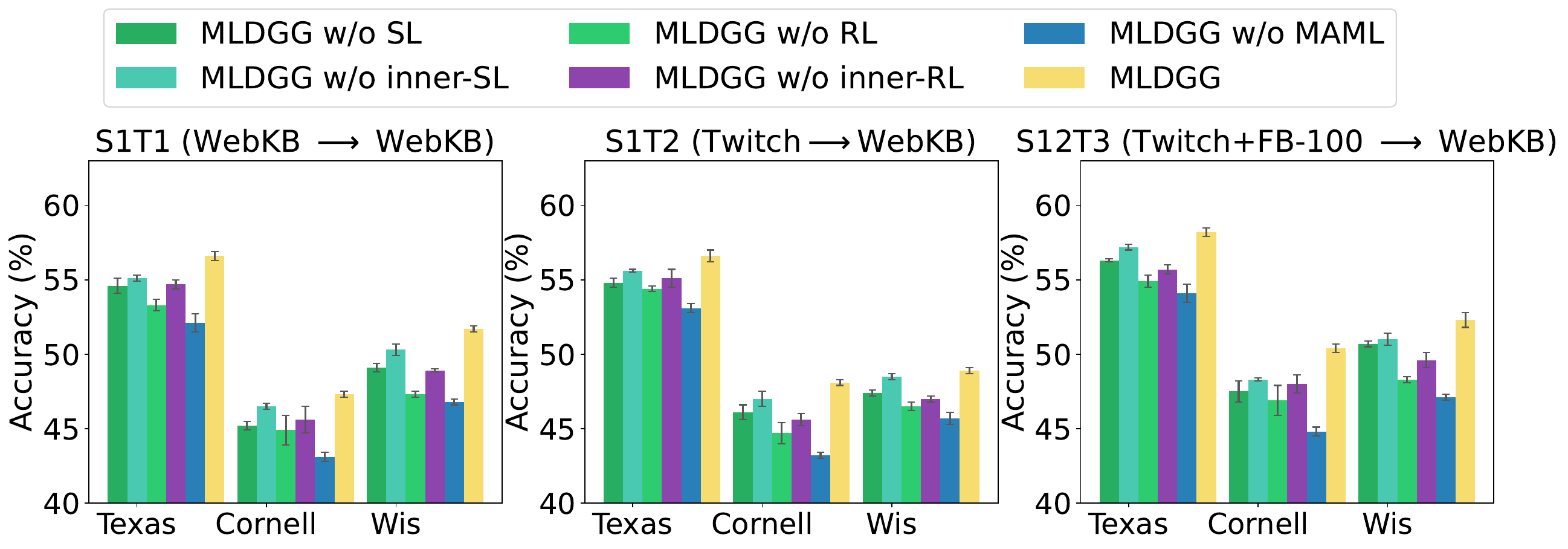}
    \caption{Ablation study for \sysname{} under three distinct cross-domain scenarios.}
	\label{fig:ablation}
    % \vspace{-3mm}
% \end{minipage}
% \vspace{-2mm}
\end{figure}

% \begin{minipage}{0.3\textwidth}
\begin{figure}[t]
    \centering
  \includegraphics[width=0.7\linewidth]{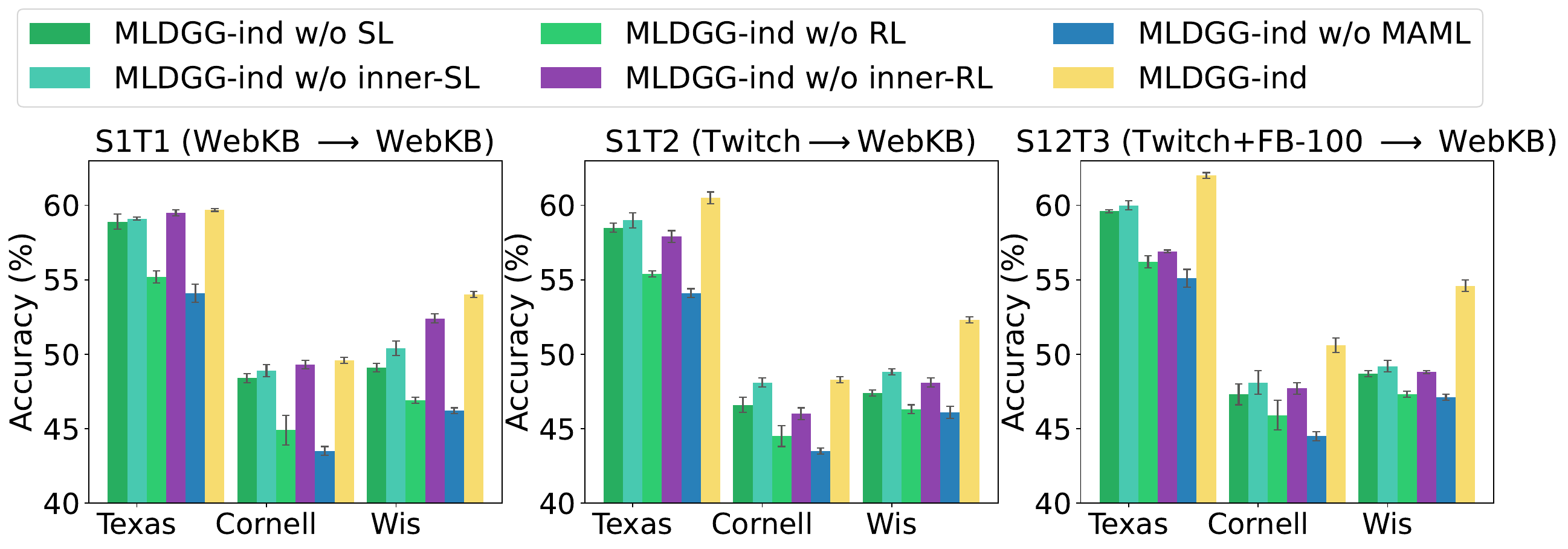}
    \caption{Ablation study for \sysname{}-ind under three distinct cross-domain scenarios.}
	\label{fig:ablation_ind}
    % \vspace{-3mm}
% \end{minipage}
 % \vspace{-1mm}
\end{figure}

\textbf{Performance Comparison.} In response to RQ1, we evaluate \sysname{}'s performance on three scenarios in \ref{Implementation}.
The results are reported in Table~\ref{tab.fb+web_twitch}-\ref{tab.cross-train-2}, where numbers in bold represent the best results and underlined means second best.
Based on these, we identified the following observations:

First, \sysname{} exhibits superior performance compared with other competitive baselines on all datasets within three experimental settings, highlighting its robust generalization capabilities.
% This phenomenon implies that meta learning across multiple domains is conducive to capturing richer knowledge among graphs and transferring knowledge across domains.
%The reason is that \sysname{} incorporates the structure and the representation learner to direct the model in improving its ability to learn enhanced node representation and semantic information recognition.
The structure learner guides the model to learn more meaningful node representations and capture shared structure information across domains, and the representation learner guides the model to disentangle semantic and domain-specific information in node representation.
Furthermore, we have observed superior classification performance when $\mathbf{s}$ and $\mathbf{v}$ are independent. 
This observation suggests that domain-specific and label-independent variation factors disturb the model's classification, consequently influencing its generalization capability. Disentangling $\mathbf{s}$ and $\mathbf{v}$ offers greater advantages for generalization.

Second, the results of S1T1, S1T2, and S12T3 demonstrate that the best generalization performance is achieved when the training domain originates from diverse datasets.
Among all comparative methods, the performance of GraphGlow is second only to our framework, which suggests that integrating structure learning with GNN facilitates the capture of richer knowledge within graphs.
All comparative methods exhibit varying performances under different distribution shift settings.
Traditional invariant learning methods exhibit superior performance in scenarios where the source domain originates from the same dataset, as opposed to situations involving diverse datasets.
It suggests that invariant learning methods demonstrate limited generalization ability in situations with substantial distribution shifts.
Other cross-domain meta-learning methods do show better generalization ability for cross-domain scenarios.
In contrast, cross-domain meta-learning methods demonstrate superior performance, attributable to meta-learning's advanced ability to capture shared knowledge across domains.

\textbf{Ablation Studies.} To answer RQ2, we conduct five ablation studies to evaluate the robustness of key modules, 
namely structure learner, representation learner, and MAML. 
The results of \sysname{} and \sysname{}-ind are shown in Fig.~\ref{fig:ablation} and Fig.~\ref{fig:ablation_ind}, where we follow the setting of Table~\ref{tab.cross-train-2}.
In-depth descriptions and the algorithms for these studies and more results can be found in Appendix~\ref{ablation_alg}.
% The full \sysname{} and \sysname{}-ind always achieve superior performance compared with the three variants, indicating the essential role of each module for domain generalization.
(1) In \textsc{\sysname{} w/o SL}, we input the graph directly into the GNN to learn node representations and compare its accuracy with \sysname{}. 
We observe declines of $2\%$ to $3\%$ in accuracy across all settings compared to the full model. 
Given that GNNs often aggregate task-irrelevant information, which can result in overfitting and diminish generalization performance, the introduction of the structure learner becomes crucial. 
By mitigating the adverse effects of task-unrelated edges, the structure learner facilitates the acquisition of comprehensive node representations, thereby improving the overall performance.
% The version without structure learner shows a larger decline in Wisconsin compared to the other.
(2) In \textsc{\sysname{} w/o RL}, we only keep the structure learner and just finetune the GNN encoder during the test phase.
% The results show that there is more loss of performance degradation.
We observe a more substantial loss in performance degradation to $3\%$ to $6\%$ across all settings. 
This indicates that the disentanglement of semantic and variation factors can enhance the model's generalization capability. Class labels are dependent on semantic factors, while variation factors representing domain-specific elements are not associated with these labels. 
When the representation learner is absent, performance degradation occurs, particularly in the presence of OOD samples stemming from distributional shifts in target domains. Therefore, mitigating the influence of variation factors becomes crucial for improving the model's robustness across diverse domains.
% This indicates that combining semantic learner with structure learner contributes more to the improvement of generalization.
(3) In \textsc{\sysname{} w/o MAML}, it does not share the semantic and variation encoders across different domains, which significantly decreases model performance by $8\%$ to $10\%$. 
This observation indicates the critical role played by the meta-learner modules in facilitating knowledge transfer from source and target graphs. 
The MAML framework serves as an integration for both the structure learner and representation learner, thereby enabling efficient knowledge transfer and facilitating effective adaptation to unseen target domains.
% Compared with the other two variants, it has the worst performance on three graphs, indicating that meta-learner modules play an important role in cross-domain knowledge transfer.
(4) In \textsc{\sysname{} w/o inner-SL}, we remove each task-specific structure learner so that all tasks share a structure learner. We observe declines of $1\%$ to $2\%$ in accuracy across all settings compared to the full model. 
This indicates that learning initialization parameters for the structure learner based on the meta-learning framework are conducive to capturing the structure information shared by different domains and improving the generalization ability.
(5) In \textsc{\sysname{} w/o inner-RL}, we remove each task-specific representation learner so that all tasks share a representation learner, which decreases model performance by $2\%$ to $3\%$.
This indicates that learning initialization parameters for the representation learner based on the meta-learning framework can guide the model to learn semantic factors and variation factors, to better disentangle to improve generalization ability.

\begin{figure}[t]
  \centering
  \includegraphics[width=0.8\linewidth]{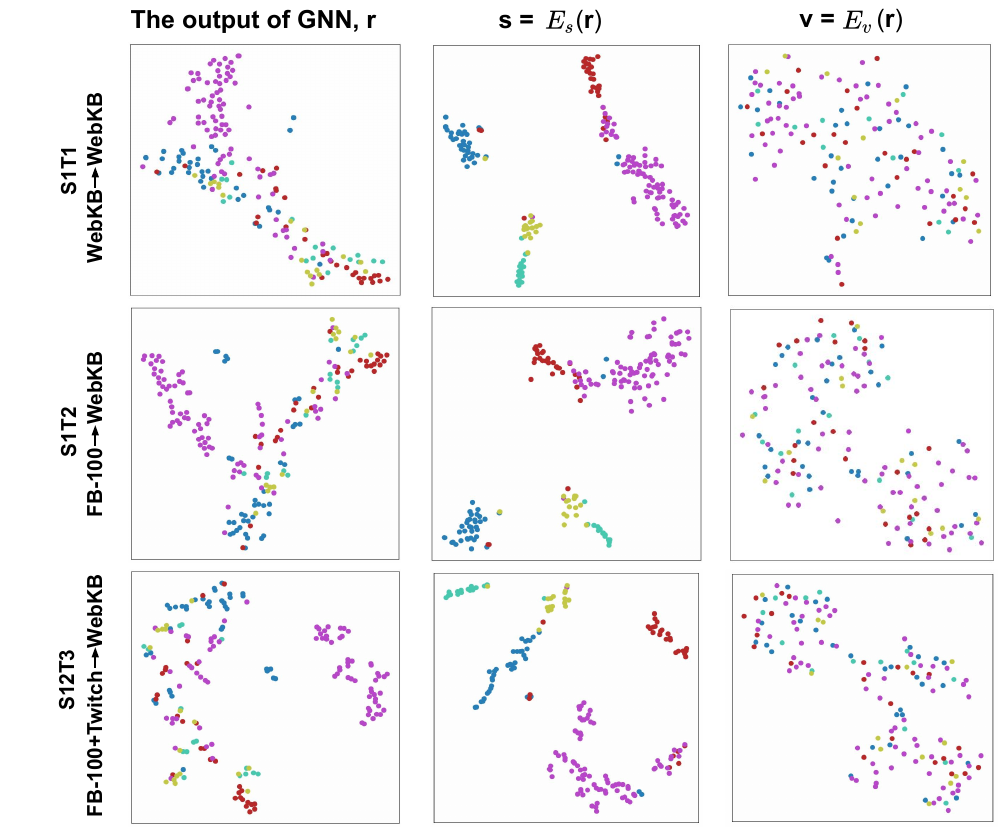}
  % \vspace{-3mm}
  \caption{ The demonstration of the effectiveness of representation learner on $3$ different setting \textbf{S12T3}, \textbf{S1T2}, and \textbf{S1T1} using T-sne visualization. $\mathbf{r}$ denotes the output of GNN, $\mathbf{s}=E_s(\mathbf{r})$ denotes the semantic factor and $\mathbf{v}=E_v(\mathbf{r})$ denotes the variation factor.
  Different colors represent different labels.
  }
  \label{fig:vis}
  % \vspace{-5mm}
\end{figure}

\textbf{The demonstration of the effectiveness of the representation learner}. 
To answer RQ3, we visualize the output $\mathbf{r}$ of each node of the GNN, domain-invariant semantic factors $\mathbf{s}$ and domain-specific variation factors 
$\mathbf{v}$ respectively in Fig.~\ref{fig:vis} (different colors represent different labels).
The domain-invariant semantic factors $\mathbf{s}$ and domain-specific variation factors $\mathbf{v}$ are disentangled from the node representations $\mathbf{r}$ learned from GNNs. 
We can see that the samples represented by $\mathbf{s}$ are more distinguished than those represented by $\mathbf{r}$. 
The samples represented by $\mathbf{v}$ are independent of classes. 
These phenomena indicate that by disentangling the node representation learned from GNN to capture the domain-invariant semantic information that determines the label, the influence of the variation factors on the label prediction can be reduced and better generalization performance can be achieved.

\textbf{Sensitivity Analysis.} The results of sensitivity analysis of \sysname{} to varying numbers of gradient steps during meta-testing and the weight of the original graph $\lambda$
are provided in Appendix~\ref{Sensitivity}.

\section{Conclusion}
In this paper, we introduce a novel cross-multi-domain meta-learning framework, \sysname{}, for node-level graph domain generalization. 
The framework integrates a structure and a representation learner within the meta-learning paradigm to facilitate knowledge transfer and enable rapid adaptation to target domains previously.
The structure learner mitigates the adverse effects of task-unrelated edges to facilitate the acquisition of comprehensive node representations of GNN while capturing the shared structure information. 
The representation learner by disentangling the semantic and variation factors enhances the model's generalization. 
% Finally, the meta-learner transfers the knowledge learned from the structure and representation learner and facilitates adaptation to target domains
% The model can train and test data originating from arbitrary distributions. 
We conduct extensive experiments 
% to assess the model's generalization capability across 
% with three distinct cross-domain settings. 
and the results demonstrate that our model outperforms baseline methods.

\clearpage
%Bibliography
\bibliographystyle{unsrt}  
\bibliography{references}  
\appendix
\clearpage

\section{Notations}
\label{notations}
For clear interpretation, we list the notations used in this paper and their corresponding explanation, as shown in Table \ref{tab:detailed_notations}.

\begin{table}[h!]
    \centering
    \caption{Important notations and corresponding descriptions.}
    \begin{tabular}{c|p{6.5cm}}
    \toprule
    \textbf{Notations} & \textbf{Descriptions} \\
        \cmidrule(lr){1-2}
         $\mathcal{G}$ &  A set of graphs\\
         $e_i$ & The $i$-th source domain\\
         $e_T$ & The target domain \\
         $G^{e_i}$ &  A graph from domain $e_i$\\
         $A^{e_i}$ & The adjacency matrix of $G^{e_i}$\\
         $X^{e_i}$ & The node feature matrix of $G^{e_i}$\\
         $\mathbf{y}^{e_i}$ & The label of $G^{e_i}$\\
         $\mathcal{V}^{e_i}$ &  A collection of nodes in $G^{e_i}$\\
         $F$ & The similarity matrix of nodes \\
         $R$ & The representation matrix of nodes in a graph $G$\\
         $A'$ & The learned adjacency matrix of structure learner\\
         $\mathcal{E}$ & The set of domains \\
         $\mathcal{E}_s$ & The set of source domains \\
         $\mathcal{E}_t$ & The set of target domains \\
         $D$ & The dimension of node feature\\
         $\mathcal{T}$ & The set of tasks in meta-learning \\
         $\mathbf{s}$ & The semantic factors\\
         $\mathbf{v}$ & The variation factors\\
         $\mathbf{r}$ & The output of GNN\\
         $E_s$ & The semantic encoder \\
         $E_v$ & The variation encoder \\
         $\hat{f} \circ g$ & A classifier \\
         $f_t$ & The structure learner \\
         $f_r$ &  The Representation learner \\
         $\boldsymbol{\theta}_{t}$ & The initialization parameters of a structure learner\\
         $\boldsymbol{\theta}_{r}$ & The initialization parameters of a representation learner\\
        $\boldsymbol{\theta}_{s}$ & The parameters of a semantic encoder\\
        $\boldsymbol{\theta}_{v}$ & The parameters of a variation encoder\\
        $\boldsymbol{\theta}_{d}$ & The parameters of a decoder\\
        $\boldsymbol{\theta}_{g}$ & The parameters of GNNs\\
        $\lambda$ &  The weight of the original graph\\
        $\lambda_r$ & The weight coefficient of  regularization loss\\
        $l_{in}$, $l_{out}$ & The inner loop and outer loop learning rate, respectively \\
        $M$ & The number of tasks\\
        $K$ & The number of source graphs\\
    \bottomrule
    \end{tabular}
    \label{tab:detailed_notations}
\end{table}

\section{Model Details}
\label{model_details}
\textbf{The Evidence Lower BOund (ELBO)}.
In this paper, we assume the representation of each node is disentangled into two factors: a domain-invariant semantic factor $\mathbf{s}$ determining the label and a domain-specific variation factor $\mathbf{v}$ independent of labels.
A common and effective approach for aligning the model 
$p$ with the data distribution $p^*(\mathbf{r},\mathbf{y})$ is through maximizing likelihood $p(\mathbf{r},\mathbf{y}) = \log p(\mathbf{r},\mathbf{y})$.
With these two latent variables, the log marginal likelihood can be $\log \int \int p(\mathbf{s},\mathbf{v},\mathbf{r}, \mathbf{y}) d\mathbf{s}d\mathbf{v}$.
However, this is an intractable problem that is difficult to evaluate and optimize.
To address this problem, one plausible way is to introduce a variational distribution $q(\mathbf{s},\mathbf{v}|\mathbf{r}, \mathbf{y})$, conditioned on the observed variables based on variational expectation-maximization (variational EM).
Then, a lower bound of the likelihood function can be derived:
\begin{equation}
        \begin{split}
          \log p(\mathbf{r},\mathbf{y}) &= \log \mathbb{E}_{q(\mathbf{s},\mathbf{v}|\mathbf{r}, \mathbf{y})} \left [\frac{p(\mathbf{s},\mathbf{v},\mathbf{r}, \mathbf{y})}{q(\mathbf{s},\mathbf{v}|\mathbf{r}, \mathbf{y})} \right] \\
          &\geqslant \mathbb{E}_{q(\mathbf{s},\mathbf{v}|\mathbf{r}, \mathbf{y})} \left [\log \frac{p(\mathbf{s},\mathbf{v},\mathbf{r}, \mathbf{y})}{q(\mathbf{s},\mathbf{v}|\mathbf{r}, \mathbf{y})} \right] \\
          &=: \mathcal{L}_{q_{\mathbf{s},\mathbf{v}|\mathbf{r}, \mathbf{y}}}(\mathbf{r},\mathbf{y})
        \end{split}
        \label{eq.ELBO_appendix}
\end{equation}
where $\mathcal{L}_{q_{\mathbf{s},\mathbf{v}|\mathbf{r}, \mathbf{y}}}(\mathbf{r},\mathbf{y})$ is called Evidence Lower BOund (ELBO).
The variational distribution $q(\mathbf{s},\mathbf{v}|\mathbf{r}, \mathbf{y})$ is commonly instantiated by a standalone model and is regarded as an inference model.
Unfortunately, the introduced model $q(\mathbf{s},\mathbf{v}|\mathbf{r}, \mathbf{y})$ fails to facilitate the estimation of $p(\mathbf{y}|\mathbf{r})$.
To alleviate this problem, we introduce an auxiliary model $q(\mathbf{s}, \mathbf{v}, \mathbf{y}|\mathbf{r})$ to target $p(\mathbf{s}, \mathbf{v}, \mathbf{y}|\mathbf{r})$, which enables the straightforward sampling of $\mathbf{y}$ given $\mathbf{r}$ for prediction. 
Meanwhile, $q(\mathbf{s}, \mathbf{v}|\mathbf{r},\mathbf{y}) = \frac{q(\mathbf{s}, \mathbf{v}, \mathbf{y}|\mathbf{r})}{q(\mathbf{y}|\mathbf{r})}$ means $q(\mathbf{s}, \mathbf{v}, \mathbf{y}|\mathbf{r})$ can help learning inference model $q(\mathbf{s}, \mathbf{v}|\mathbf{r},\mathbf{y})$, where $q(\mathbf{y}|\mathbf{r}) := \int q(\mathbf{s}, \mathbf{v}, \mathbf{y}|\mathbf{r})d\mathbf{s}d\mathbf{v}$.
When the ELBO approaches its maximum, all posterior items gradually tend to converge towards the prior, thus $q(\mathbf{s}, \mathbf{v}, \mathbf{y}|\mathbf{r}) = p(\mathbf{s}, \mathbf{v}, \mathbf{y}|\mathbf{r}) = p(\mathbf{s}, \mathbf{v}| \mathbf{r}) p(\mathbf{y}| \mathbf{s})$.
For $p(\mathbf{s}, \mathbf{v}|\mathbf{r})$, we instead use inference model $q(\mathbf{s}, \mathbf{v}|\mathbf{r})$.
Then, $q(\mathbf{s}, \mathbf{v}| \mathbf{y},\mathbf{r}) = \frac{q(\mathbf{s}, \mathbf{v} | \mathbf{r})p(\mathbf{y}|\mathbf{s})}{q(\mathbf{y}| \mathbf{r})}$. Then, the ELBO is turned to:
\begin{equation}
% \footnotesize
\begin{split}
    & \mathbb{E}_{q(\mathbf{s},\mathbf{v}|\mathbf{r}, \mathbf{y})} \left [\log \frac{p(\mathbf{s},\mathbf{v},\mathbf{r}, \mathbf{y})}{q(\mathbf{s},\mathbf{v}|\mathbf{r}, \mathbf{y})} \right] \\
    &=\mathbb{E}_{q(\mathbf{s},\mathbf{v}|\mathbf{r}, \mathbf{y})} \left [\log \frac{p(\mathbf{s},\mathbf{v},\mathbf{r}, \mathbf{y}) q(\mathbf{y}|\mathbf{r})}{q(\mathbf{s},\mathbf{v}|\mathbf{r}) p(\mathbf{y}|\mathbf{s})} \right] \\
    &=\mathbb{E}_{q(\mathbf{s},\mathbf{v}|\mathbf{r}, \mathbf{y})} \left [\log \frac{p(\mathbf{s},\mathbf{v}) p(\mathbf{r}|\mathbf{s},\mathbf{v}) p(\mathbf{y}|\mathbf{s})q(\mathbf{y}|\mathbf{r})}{q(\mathbf{s},\mathbf{v}|\mathbf{r}) p(\mathbf{y}|\mathbf{s})} \right] \\
    & = \mathbb{E}_{\frac{q(\mathbf{s},\mathbf{v}|\mathbf{r})p(\mathbf{y}|\mathbf{s})}{q(\mathbf{y}|\mathbf{r})}} \left [\log \frac{p(\mathbf{s},\mathbf{v}) p(\mathbf{r}|\mathbf{s},\mathbf{v}) p(\mathbf{y}|\mathbf{s})q(\mathbf{y}|\mathbf{r})}{q(\mathbf{s},\mathbf{v}|\mathbf{r}) p(\mathbf{y}|\mathbf{s})} \right] \\
    & = \mathbb{E}_{\frac{q(\mathbf{s},\mathbf{v}|\mathbf{r})p(\mathbf{y}|\mathbf{s})}{q(\mathbf{y}|\mathbf{r})}} \left [\log \frac{p(\mathbf{s},\mathbf{v}) p(\mathbf{r}|\mathbf{s},\mathbf{v})}{q(\mathbf{s},\mathbf{v}|\mathbf{r})} + \log q(\mathbf{y}|\mathbf{r})\right] \\
    &= \mathbb{E}_{q(\mathbf{s},\mathbf{v}|\mathbf{r})p(\mathbf{y}|\mathbf{s})} \left [\frac{1}{q(\mathbf{y}|\mathbf{r})} \log q(\mathbf{y}|\mathbf{r})\right] + \mathbb{E}_{q(\mathbf{s},\mathbf{v}|\mathbf{r})p(\mathbf{y}|\mathbf{s})}\left [\frac{1}{q(\mathbf{y}|\mathbf{r})} \log \frac{p(\mathbf{s},\mathbf{v}) p(\mathbf{r}|\mathbf{s},\mathbf{v})}{q(\mathbf{s},\mathbf{v}|\mathbf{r})} \right] \\
    &= \mathbb{E}_{q(\mathbf{s},\mathbf{v}|\mathbf{r})} \left [\frac{p(\mathbf{y}|\mathbf{s})\log q(\mathbf{y}|\mathbf{r})}{q(\mathbf{y}|\mathbf{r})}  \right] + \mathbb{E}_{q(\mathbf{s},\mathbf{v}|\mathbf{r})}\left [\frac{p(\mathbf{y}|\mathbf{s})}{ q(\mathbf{y}|\mathbf{r})} \left(\log p(\mathbf{r} | \mathbf{s},\mathbf{v}) + \log \frac{p(\mathbf{s},\mathbf{v})}{q(\mathbf{s},\mathbf{v}|\mathbf{r})} \right)\right] \\
    &=\frac{1}{q(\mathbf{y}|\mathbf{r})} \left[ \mathbb{E}_{q(\mathbf{s},\mathbf{v}|\mathbf{r})}\left [p(\mathbf{y}|\mathbf{s}) \log q(\mathbf{y} | \mathbf{r})\right] + \mathbb{E}_{q(\mathbf{s},\mathbf{v}|\mathbf{r})}\left[p(\mathbf{y}|\mathbf{s}) \log p(\mathbf{r} | \mathbf{s},\mathbf{v})\right] \right]\\
    &+  \frac{1}{q(\mathbf{y}|\mathbf{r})} \left[\mathbb{E}_{q(\mathbf{s},\mathbf{v}|\mathbf{r})}\left[p(\mathbf{y}|\mathbf{s}) \log \frac{p(\mathbf{s}, \mathbf{v})}{q(\mathbf{s},\mathbf{v}|\mathbf{r})} \right] \right]\\
    \end{split}
    \label{eq.elbo_inference}
\end{equation}
where $q(\mathbf{y}|\mathbf{r}) = \mathbb{E}_{q(\mathbf{s},\mathbf{v}|\mathbf{r})} \left [ p(\mathbf{y}|\mathbf{s})\right]$.

\subsection{Complexity Analysis}
\label{complexity}
In our experiments followed by \cite{zhao2023graphglow}, the complexity of the structure learner is $O(NP)$, where $P$ is is the number of pivot nodes.
The complexity of GCNs is $O(|E|Dd)$, where $|E|$ and $d$ are the number of edges and classed, $D$ is is the dimension of the node feature, respectively.
The complexity of the representation learner is $O(N)$.
Therefore, the complexity of our model is $O(K(NP+\eta(|E|Dd+N)))$, where $K$ is the number of source domains and $K,P,\eta \ll N$, and $D,d \ll |E|$.

\section{Datasets And Experimental Details}
\subsection{Datasets}
\label{dataset}
\begin{itemize}[leftmargin=*]
    \item \textsc{Twitch-Explicit}~\cite{rozemberczki2021multi}. It is a gamer network that includes seven networks: DE, ENGB, ES, FR, PTBR, RU, and TW.
    Each network represents a particular game region. 
    The aforementioned networks have comparable sizes but vary in terms of densities and maximum node degrees. 
    \item \textsc{Facebook-100}~\cite{traud2012social}. This dataset comprises 100 snapshots of the \textit{Facebook} friendship network, dating back to 2005.
    Each node represents a user from a particular American university, and the edges indicate the friendships between these users.
    We use five networks in our experiments: Amherst, John Hopkins, Reed98, Cornell5, and Yale4.
    \item \textsc{WebKB}~\cite{pei2020geom}. It is a web page network dataset. 
    The nodes in the network represent web pages, and the edges symbolize hyperlinks connecting these web pages. Additionally, the node features are represented using the bag-of-words representation of the web pages.
    The task is to classify the nodes into one of five categories: student, project, course, staff, and faculty. 
    According to the university. it is split into three networks: Cornell, Texas, and Wisconsin.
    
\end{itemize}

\subsection{Baseline}
\label{baseline}
\begin{itemize}[leftmargin=*]
\item \textbf{EERM}~\cite{wu2022handling} generates environments with diverse topologies and then minimizes the variances and mean values of predicted loss across different environments.
\item \textbf{SRGNN}~\cite{zhu2021shift} is devoted to solving the distributional shift problem by converting biased data sets to unbiased data distribution.
\item \textbf{Mixup}~\cite{zhang2017mixup} improves model generation capacity by constructing novel training examples drawn from raw data, thereby expanding the training distribution.
\item \textbf{GraphGlow}~\cite{zhao2023graphglow}  employs a meta-learning approach to cultivate a generalized structure learner aimed at discerning universally applicable patterns in optimal messaging topologies across diverse datasets.
\item \textbf{GMeta}~\cite{huang2020graph} exhibits the capacity to generalize to Graph Neural Networks (GNNs) applied to entirely new graphs and labels that have not been encountered previously. Simultaneously, it showcases the ability to find evidence supporting predictions based on small datasets within local subgraphs surrounding target nodes or edges.
\item \textbf{FLOOD}~\cite{liu2023flood} employs an adaptive encoder, refined through invariant learning and bootstrapped learning strategies, to enhance performance on a test set. First, it constructs a shared encoder by minimizing the empirical risk across various domains. Then, it utilizes bootstrapped learning with a self-supervised method to tailor the shared encoder for optimal adaptation to the test set.
\item \textbf{MD-Gram}~\cite{lin2023multi} is a multi-domain graph meta-learning approach, transforming learning tasks from multiple source-domain graphs into a unified domain. This process facilitates the acquisition of transferable knowledge across domains.
\end{itemize}

\section{Ablation Studies}
\label{ablation_alg}
We conduct five ablation studies, and detailed algorithms of designed ablation studies are given in Algorithms~\ref{algorithm_wo_SL} to \ref{algorithm_wo_innerRL}.

\begin{algorithm}[t!]
    \caption{\textbf{ \sysname{} W/O SL (Ablation Study 1)}} 
    \label{algorithm_wo_SL}
    \begin{algorithmic}[1]
    \STATE \textbf{While} not done \textbf{do}: 
    \STATE \qquad \textbf{For} each source graph $\mathcal{T}^i$ \textbf{do}:
    \STATE \qquad \qquad $R_0^i =X^i$
    \STATE \qquad \qquad $\boldsymbol{\theta}'^i_{s_{0}} = \boldsymbol{\theta}_s$, $\boldsymbol{\theta}'^i_{v_{0}} = \boldsymbol{\theta}_v$, $\boldsymbol{\theta}'^i_{d_{0}} = \boldsymbol{\theta}_d$,$\boldsymbol{\theta}^i_{g_{0}} = \boldsymbol{\theta}_g^i$
    \STATE \qquad \qquad Sample  $\mathcal{T}_{qry}^{i}$ and $\mathcal{T}_{sup}^{i}$
     \STATE \qquad \qquad \textbf{For} n in $1, \dots, \eta$ \textbf{do}:
     \STATE \qquad \qquad \qquad  Compute $\mathcal{L}_{sup}^i$ on $\mathcal{T}_{sup}^{i}$ via Eq.~\eqref{eq.loss_total}
     \STATE \qquad \qquad \qquad
     $\boldsymbol{\theta}'^i_{s_{n}} = \boldsymbol{\theta}'^i_{s_{{n-1}}} - l_{in} \bigtriangledown \mathcal{L}_{sup}^i$
      \STATE \qquad \qquad \qquad
     $\boldsymbol{\theta}'^i_{v_{n}} = \boldsymbol{\theta}'^i_{v_{{n-1}}} - l_{in} \bigtriangledown \mathcal{L}_{sup}^i$
     \STATE \qquad \qquad \qquad
     $\boldsymbol{\theta}'^i_{d_{n}} = \boldsymbol{\theta}'^i_{d_{{n-1}}} - l_{in} \bigtriangledown \mathcal{L}_{sup}^i$
      \STATE \qquad \qquad \qquad
     $\boldsymbol{\theta}_{g_n}^i = \boldsymbol{\theta}^i_{g_{{n-1}}} - l_{in} \bigtriangledown \mathcal{L}_{sup}^i$
     \STATE \qquad \qquad \qquad Compute $\mathcal{L}_{qry}^{i,n}$ on $\mathcal{T}_{qry}^{i}$ via Eq.~\eqref{eq.loss_total}
     \STATE \qquad \qquad \textbf{End}
     \STATE \qquad \qquad $\mathcal{L}_{qry}^{i} = \mathcal{L}_{qry}^{i,\eta}$
     \STATE \qquad \textbf{End}
    \STATE \qquad Update $\boldsymbol{\theta}_s \leftarrow \boldsymbol{\theta}_s - l_{out} \bigtriangledown_{\boldsymbol{\theta}_s} \frac{1}{M} \sum_{i=1}^M \mathcal{L}_{qry}^{i} $
    \STATE \qquad Update $\boldsymbol{\theta}_v \leftarrow \boldsymbol{\theta}_v - l_{out} \bigtriangledown_{\boldsymbol{\theta}_v} \frac{1}{M} \sum_{i=1}^M \mathcal{L}_{qry}^{i}$
    \STATE \qquad Update $\boldsymbol{\theta}_d \leftarrow \boldsymbol{\theta}_d - l_{out} \bigtriangledown_{\boldsymbol{\theta}_d} \frac{1}{M} \sum_{i=1}^M \mathcal{L}_{qry}^{i}$
    \STATE \textbf{End} while
    \end{algorithmic}
\end{algorithm}

\begin{algorithm}[t!]
    \caption{\textbf{ \sysname{} W/O RL (Ablation Study 2)}} \label{algorithm_wo_RL}
    \begin{algorithmic}[1]
      \STATE \textbf{While} not done \textbf{do}: 
    \STATE \qquad \textbf{For} each task $\mathcal{T}^i$ \textbf{do}:
    \STATE \qquad \qquad  $R_0^i =X^i$
    \STATE \qquad \qquad  Compute $F^i$ using Eq.~\eqref{eq.edge_weight}
    \STATE \qquad  \qquad Sample $H$ times over $F^i$ to obtain  $A'^i$
    \STATE \qquad \qquad 
    Compute $R^i$ using Eq.~\eqref{eq.representation}
    % \STATE \qquad \qquad  $\{ \mathbf{r}_{sup}, \mathbf{y}_{sup} \}$, $\{ \mathbf{r}_{qry}, \mathbf{y}_{qry} \}$
    \STATE \qquad \qquad $\boldsymbol{\theta}'^i_{t_{0}} = \boldsymbol{\theta}_t$, $\boldsymbol{\theta}^i_{g_{0}} = \boldsymbol{\theta}_g^i$
    \STATE \qquad \qquad Sample  $\mathcal{T}_{qry}^{i}$ and $\mathcal{T}_{sup}^{i}$
    \STATE \qquad \qquad \textbf{For} n in $1, \dots, \eta$ \textbf{do}:
     \STATE \qquad \qquad \qquad  Compute $\mathcal{L}_{sup}^i$ on $\mathcal{T}_{sup}^{i}$ via Eq.~\eqref{eq.loss_total}
    \STATE \qquad \qquad \qquad
     $\boldsymbol{\theta}'^i_{t_{n}} = \boldsymbol{\theta}'^i_{t_{{n-1}}} - l_{in}\bigtriangledown \mathcal{L}_{sup}^i$
      \STATE \qquad \qquad \qquad
     $\boldsymbol{\theta}_{g_n}^i = \boldsymbol{\theta}^i_{g_{{n-1}}} - l_{in} \bigtriangledown \mathcal{L}_{sup}^i$
     \STATE \qquad \qquad \qquad Compute $\mathcal{L}_{qry}^{i,n}$ on $\mathcal{T}_{qry}^{i}$ via Eq.~\eqref{eq.loss_total}
     \STATE \qquad \qquad \textbf{End}
     \STATE \qquad \qquad $\mathcal{L}_{qry}^{i} = \mathcal{L}_{qry}^{i,\eta}$
     \STATE \qquad \textbf{End}
    \STATE \qquad Update $\boldsymbol{\theta}_t \leftarrow \boldsymbol{\theta}_t - l_{out} \bigtriangledown_{\boldsymbol{\theta}_t} \frac{1}{M} \sum_{i=1}^M \mathcal{L}^i_{qry}$
    \STATE \textbf{End} while
    \end{algorithmic}
\end{algorithm}

\begin{algorithm}[t!]
    \caption{\textbf{ \sysname{} W/O MAML (Ablation Study 3)}} \label{algorithm_wo_MAML}
    \begin{algorithmic}[1]
        \STATE \textbf{While} not done \textbf{do}: 
    \STATE \qquad \textbf{For} each source graphs $G^i$ \textbf{do}:
    \STATE \qquad \qquad  $R_0^i =X^i$
    \STATE \qquad \qquad  Compute $F^i$ using Eq.~\eqref{eq.edge_weight}
    \STATE \qquad  \qquad Sample $H$ times over $F^i$ to obtain  $A'^i$
    \STATE \qquad \qquad 
    Compute $R^i$ using Eq.~\eqref{eq.representation}
    % \STATE \qquad \qquad  $\{ \mathbf{r}_{sup}, \mathbf{y}_{sup} \}$, $\{ \mathbf{r}_{qry}, \mathbf{y}_{qry} \}$
    \STATE \qquad \qquad $\boldsymbol{\theta}'^i_{t_{0}} = \boldsymbol{\theta}_t$, $\boldsymbol{\theta}'^i_{s_{0}} = \boldsymbol{\theta}_s$, $\boldsymbol{\theta}'^i_{v_{0}} = \boldsymbol{\theta}_v$, $\boldsymbol{\theta}'^i_{d_{0}} = \boldsymbol{\theta}_d$,$\boldsymbol{\theta}^i_{g_{0}} = \boldsymbol{\theta}_g^i$
    \STATE \qquad \qquad \textbf{For} n in $1, \dots, \eta$ \textbf{do}:
     \STATE \qquad \qquad \qquad  Compute $\mathcal{L}^i$ via Eq.~\eqref{eq.loss_total}
    \STATE \qquad \qquad \qquad
     $\boldsymbol{\theta}'^i_{t_{n}} = \boldsymbol{\theta}'^i_{t_{{n-1}}} - l_{in} \bigtriangledown \mathcal{L}^i$
     \STATE \qquad \qquad \qquad
     $\boldsymbol{\theta}'^i_{s_{n}} = \boldsymbol{\theta}'^i_{s_{{n-1}}} - l_{in} \bigtriangledown \mathcal{L}^i$
      \STATE \qquad \qquad \qquad
     $\boldsymbol{\theta}'^i_{v_{n}} = \boldsymbol{\theta}'^i_{v_{{n-1}}} - l_{in} \bigtriangledown \mathcal{L}^i$
     \STATE \qquad \qquad \qquad
     $\boldsymbol{\theta}'^i_{d_{n}} = \boldsymbol{\theta}'^i_{d_{{n-1}}} - l_{in} \bigtriangledown \mathcal{L}^i$
      \STATE \qquad \qquad \qquad
     $\boldsymbol{\theta}_{g_n}^i = \boldsymbol{\theta}^i_{g_{{n-1}}} - l_{in} \bigtriangledown \mathcal{L}^i$
     \STATE \qquad \qquad \textbf{End}
     \STATE \qquad \textbf{End}
    \STATE \textbf{End} while
    \end{algorithmic}
\end{algorithm}

\begin{algorithm}[t!]
    \caption{\textbf{ \sysname{} W/O inner-SL (Ablation Study 4)}} 
    \label{algorithm_wo_innerSL}
    \begin{algorithmic}[1]
        \STATE \textbf{While} not done \textbf{do}: 
    \STATE \qquad \textbf{For} each task $\mathcal{T}^i$ \textbf{do}:
    \STATE \qquad \qquad  $R_0^i =X^i$
    \STATE \qquad \qquad  Compute $F^i$ using Eq.~\eqref{eq.edge_weight};
    \STATE \qquad  \qquad Sample $H$ times over $F^i$ to obtain  $A'^i$
    \STATE \qquad \qquad 
    Compute $R^i$ using Eq.~\eqref{eq.representation}
    % \STATE \qquad \qquad  $\{ \mathbf{r}_{sup}, \mathbf{y}_{sup} \}$, $\{ \mathbf{r}_{qry}, \mathbf{y}_{qry} \}$
    \STATE \qquad \qquad $\boldsymbol{\theta}'^i_{t_{0}} = \boldsymbol{\theta}_t$, $\boldsymbol{\theta}'^i_{s_{0}} = \boldsymbol{\theta}_s$, $\boldsymbol{\theta}'^i_{v_{0}} = \boldsymbol{\theta}_v$, $\boldsymbol{\theta}'^i_{d_{0}} = \boldsymbol{\theta}_d$,$\boldsymbol{\theta}^i_{g_{0}} = \boldsymbol{\theta}_g^i$
    \STATE \qquad \qquad Sample  $\mathcal{T}_{qry}^{i}$ and $\mathcal{T}_{sup}^{i}$
    \STATE \qquad \qquad \textbf{For} n in $1, \dots, \eta$ \textbf{do}:
     \STATE \qquad \qquad \qquad  Compute $\mathcal{L}_{sup}^i$ on $\mathcal{T}_{sup}^{i}$ via Eq.~\eqref{eq.loss_total}
     \STATE \qquad \qquad \qquad
     $\boldsymbol{\theta}'^i_{s_{n}} = \boldsymbol{\theta}'^i_{s_{{n-1}}} - l_{in} \bigtriangledown \mathcal{L}_{sup}^i$
      \STATE \qquad \qquad \qquad
     $\boldsymbol{\theta}'^i_{v_{n}} = \boldsymbol{\theta}'^i_{v_{{n-1}}} - l_{in} \bigtriangledown \mathcal{L}_{sup}^i$
     \STATE \qquad \qquad \qquad
     $\boldsymbol{\theta}'^i_{d_{n}} = \boldsymbol{\theta}'^i_{d_{{n-1}}} - l_{in} \bigtriangledown \mathcal{L}_{sup}^i$
      \STATE \qquad \qquad \qquad
     $\boldsymbol{\theta}_{g_n}^i = \boldsymbol{\theta}^i_{g_{{n-1}}} - l_{in} \bigtriangledown \mathcal{L}_{sup}^i$
     \STATE \qquad \qquad \qquad Compute $\mathcal{L}_{qry}^{i,n}$ on $\mathcal{T}_{qry}^{i}$ via Eq.~\eqref{eq.loss_total}
     \STATE \qquad \qquad \textbf{End}
     \STATE \qquad \qquad $\mathcal{L}_{qry}^{i} = \mathcal{L}_{qry}^{i,\eta}$
     \STATE \qquad \textbf{End}
    \STATE \qquad Update $\boldsymbol{\theta}_t \leftarrow \boldsymbol{\theta}_t - l_{out} \bigtriangledown_{\boldsymbol{\theta}_t} \frac{1}{M} \sum_{i=1}^M \mathcal{L}^i_{qry}$
    \STATE \qquad Update $\boldsymbol{\theta}_s \leftarrow \boldsymbol{\theta}_s - l_{out} \bigtriangledown_{\boldsymbol{\theta}_s} \frac{1}{M} \sum_{i=1}^M \mathcal{L}_{qry}^{i} $
    \STATE \qquad Update $\boldsymbol{\theta}_v \leftarrow \boldsymbol{\theta}_v - l_{out} \bigtriangledown_{\boldsymbol{\theta}_v} \frac{1}{M} \sum_{i=1}^M \mathcal{L}_{qry}^{i}$
    \STATE \qquad Update $\boldsymbol{\theta}_d \leftarrow \boldsymbol{\theta}_d - l_{out} \bigtriangledown_{\boldsymbol{\theta}_d} \frac{1}{M} \sum_{i=1}^M \mathcal{L}_{qry}^{i}$
    \STATE \textbf{End} while
    \end{algorithmic}
\end{algorithm}

\begin{algorithm}[t!]
    \caption{\textbf{ \sysname{} W/O inner-RL (Ablation Study 5)}} 
    \label{algorithm_wo_innerRL}
    \begin{algorithmic}[1]
     \STATE \textbf{While} not done \textbf{do}: 
    \STATE \qquad \textbf{For} each task $\mathcal{T}^i$ \textbf{do}:
    \STATE \qquad \qquad  $R_0^i =X^i$
    \STATE \qquad \qquad  Compute $F^i$ using Eq.~\eqref{eq.edge_weight}
    \STATE \qquad  \qquad Sample $H$ times over $F^i$ to obtain  $A'^i$
    \STATE \qquad \qquad 
    Compute $R^i$ using Eq.~\eqref{eq.representation}
    % \STATE \qquad \qquad  $\{ \mathbf{r}_{sup}, \mathbf{y}_{sup} \}$, $\{ \mathbf{r}_{qry}, \mathbf{y}_{qry} \}$
    \STATE \qquad \qquad $\boldsymbol{\theta}'^i_{t_{0}} = \boldsymbol{\theta}_t$, $\boldsymbol{\theta}'^i_{s_{0}} = \boldsymbol{\theta}_s$, $\boldsymbol{\theta}'^i_{v_{0}} = \boldsymbol{\theta}_v$, $\boldsymbol{\theta}'^i_{d_{0}} = \boldsymbol{\theta}_d$,$\boldsymbol{\theta}^i_{g_{0}} = \boldsymbol{\theta}_g^i$
    \STATE \qquad \qquad Sample  $\mathcal{T}_{qry}^{i}$ and $\mathcal{T}_{sup}^{i}$
    \STATE \qquad \qquad \textbf{For} n in $1, \dots, \eta$ \textbf{do}:
     \STATE \qquad \qquad \qquad  Compute $\mathcal{L}_{sup}^i$ on $\mathcal{T}_{sup}^{i}$ via Eq.~\eqref{eq.loss_total}
    \STATE \qquad \qquad \qquad
     $\boldsymbol{\theta}'^i_{t_{n}} = \boldsymbol{\theta}'^i_{t_{{n-1}}} -l_{in} \bigtriangledown \mathcal{L}_{sup}^i$
     \STATE \qquad \qquad \qquad
     $\boldsymbol{\theta}_{g_n}^i = \boldsymbol{\theta}^i_{g_{{n-1}}} - l_{in} \bigtriangledown \mathcal{L}_{sup}^i$
     \STATE \qquad \qquad \qquad Compute $\mathcal{L}_{qry}^{i,n}$ on $\mathcal{T}_{qry}^{i}$ via Eq.~\eqref{eq.loss_total}
     \STATE \qquad \qquad \textbf{End}
     \STATE \qquad \qquad $\mathcal{L}_{qry}^{i} = \mathcal{L}_{qry}^{i,\eta}$
     \STATE \qquad \textbf{End}
    \STATE \qquad Update $\boldsymbol{\theta}_t \leftarrow \boldsymbol{\theta}_t - l_{out} \bigtriangledown_{\boldsymbol{\theta}_t} \frac{1}{M} \sum_{i=1}^M \mathcal{L}^i_{qry}$
    \STATE \qquad Update $\boldsymbol{\theta}_s \leftarrow \boldsymbol{\theta}_s - l_{out} \bigtriangledown_{\boldsymbol{\theta}_s} \frac{1}{M} \sum_{i=1}^M \mathcal{L}_{qry}^{i} $
    \STATE \qquad Update $\boldsymbol{\theta}_v \leftarrow \boldsymbol{\theta}_v - l_{out} \bigtriangledown_{\boldsymbol{\theta}_v} \frac{1}{M} \sum_{i=1}^M \mathcal{L}_{qry}^{i}$
    \STATE \qquad Update $\boldsymbol{\theta}_d \leftarrow \boldsymbol{\theta}_d - l_{out} \bigtriangledown_{\boldsymbol{\theta}_d} \frac{1}{M} \sum_{i=1}^M \mathcal{L}_{qry}^{i}$
    \STATE \textbf{End} while
    \end{algorithmic}
\end{algorithm}

% \subsection{More Experimental Results}
% \label{result}

\section{Theoretical Gurantee}
\label{proof}
\textbf{JS distance.}
We adopt Jensen-Shannon (JS) distance \cite{endres2003new} to measure the dissimilarity between two distributions. Formally, JS distance between distributions $\mathbb{P}$ and $\mathbb{P}'$ is defined as $$d_{JS}(\mathbb{P},\mathbb{P}'):=\sqrt{\mathcal{D}_{JS}(\mathbb{P}||\mathbb{P}')},$$
where $\mathcal{D}_{JS}(\mathbb{P}||\mathbb{P}'):= \frac{1}{2}\mathcal{D}_{KL}(\mathbb{P}||\frac{\mathbb{P}+\mathbb{P}'}{2})+\frac{1}{2}\mathcal{D}_{KL}(\mathbb{P}'||\frac{\mathbb{P}+\mathbb{P}'}{2})$ is JS divergence defined based on Kullback–Leibler (KL) divergence $\mathcal{D}_{KL}(\cdot||\cdot)$. Note that, unlike KL divergence, JS divergence is symmetric and bounded: $0\leq \mathcal{D}_{JS}(\mathbb{P}||\mathbb{P}')\leq 1$.

\subsection{Proof for Theorem~\ref{theorem-1}}
\begin{proof}
Taking the average of upper bounds based on all source domains, we can have:

\begin{align}
% \footnotesize
\epsilon_{\texttt{Acc}}^{e_{T}}\left(\hat{f} \circ g \right) &\leq \frac{1}{K}\sum_{i=1}^{K}\epsilon_{\texttt{Acc}}^{e_{i}}\left(\hat{f} \circ g \right) + \frac{\sqrt{2} \pi_u}{K} \sum_{i=1}^{K}d_{JS}\left(\mathbb{P}^{e_{T}}_{S,Y}, \mathbb{P}^{e_{i}}_{S,Y} \right) \nonumber\\
    &\overset{(1)}{\leq} \frac{1}{K}\sum_{i=1}^{K}\epsilon_{\texttt{Acc}}^{e_{i}}\left(\hat{f} \circ g \right) + \frac{\sqrt{2} \pi_u}{K} \sum_{i=1}^{K} d_{JS}\left(\mathbb{P}^{e_{T}}_{S,Y}, \mathbb{P}^{e_{*}}_{S,Y} \right) \nonumber\\&\quad+ \frac{\sqrt{2} \pi_u}{K} \sum_{i=1}^{K} d_{JS}\left(\mathbb{P}^{e_{*}}_{S,Y}, \mathbb{P}^{e_{i}}_{S,Y} \right) \nonumber\\
    &\overset{(2)}{\leq} \frac{1}{K}\sum_{i=1}^{K}\epsilon_{\texttt{Acc}}^{e_{i}}\left(\hat{f} \circ g \right) + \sqrt{2} \pi_u \underset{i \in [K]}{\min} d_{JS}\left(\mathbb{P}^{e_{T}}_{S,Y}, \mathbb{P}^{e_{i}}_{S,Y} \right) \nonumber\\&\quad+ \sqrt{2} \pi_u \underset{i, j \in [K]}{\max} d_{JS}\left(\mathbb{P}^{e_{i}}_{S,Y}, \mathbb{P}^{e_{j}}_{S,Y} \right) 
    \label{eq-s1}
\end{align}
Here we have $\overset{(1)}{\leq}$ by using triangle inequality for JS-distance: $d_{JS}(\mathbb{P}, \mathbb{Z}) \leq d_{JS}(\mathbb{P}, \mathbb{Q}) + d_{JS}(\mathbb{Q}, \mathbb{P})$ with $\mathbb{P}, \mathbb{Q},$ and $\mathbb{Z} = \mathbb{P}^{e_{T}}, \mathbb{P}^{e_{*}}$ and $\mathbb{P}^{e_{i}}$, respectively. The previous work have $\overset{(2)}{\leq}$ because $e_{*} \in \{e_{i} \}_{i=1}^{K}$ then $d_{JS}\left(\mathbb{P}^{e_{*}}_{S,Y}, \mathbb{P}^{e_{i}}_{S,Y} \right) \leq  \underset{i, j \in [K]}{\max} d_{JS}\left(\mathbb{P}^{e_{i}}_{S,Y}, \mathbb{P}^{e_{j}}_{S,Y} \right)$. However, upon examination, we believe that using the min symbol for the second term on the right side of $\overset{(2)}{\leq}$ is not rigorous. Here, we have corrected it to use the max symbol. Therefore, we have 
\begin{align}
% \footnotesize
\epsilon_{\texttt{Acc}}^{e_{T}}\left(\hat{f} \circ g \right) &\leq \frac{1}{K}\sum_{i=1}^{K}\epsilon_{\texttt{Acc}}^{e_{i}}\left(\hat{f} \circ g \right) + \frac{\sqrt{2} \pi_u}{K} \sum_{i=1}^{K}d_{JS}\left(\mathbb{P}^{e_{T}}_{S,Y}, \mathbb{P}^{e_{i}}_{S,Y} \right) \nonumber\\
    &{\leq} \frac{1}{N}\sum_{i=1}^{K}\epsilon_{\texttt{Acc}}^{e_{i}}\left(\hat{f} \circ g \right) + \sqrt{2} \pi_u \underset{i \in [K]}{\max} d_{JS}\left(\mathbb{P}^{e_{T}}_{S,Y}, \mathbb{P}^{e_{i}}_{S,Y} \right) \nonumber\\&\quad+ \sqrt{2} \pi_u \underset{i, j \in [K]}{\max} d_{JS}\left(\mathbb{P}^{e_{i}}_{S,Y}, \mathbb{P}^{e_{j}}_{S,Y} \right).
    \label{eq-s1}
\end{align}
Similarly, we can obtain the upper bound based on the feature space $\mathcal{A} \times \mathcal{X}$ as follows:
\begin{align}
% \footnotesize
    \epsilon_{\texttt{Acc}}^{e_{T}}\left(\hat{f} \circ g \right)
    &\leq \frac{1}{K}\sum_{i=1}^{N}\epsilon_{\texttt{Acc}}^{e_{i}}\left(\hat{f} \circ g \right) + \sqrt{2} \pi_u \underset{i \in [K]}{\max} d_{JS}\left(\mathbb{P}^{e_{T}}_{A,X,Y}, \mathbb{P}^{e_{i}}_{A,X,Y} \right) \nonumber\\&\quad+ \sqrt{2} \pi_u \underset{i, j \in [K]}{\max} d_{JS}\left(\mathbb{P}^{e_{i}}_{A,X,Y}, \mathbb{P}^{e_{j}}_{A,X,Y} \right) \label{eq-s2}.
\end{align}
\begin{lemma}\label{lemma:1}
Consider two distributions $\mathbb{P}^{e_{i}}_{A,X}$ and $\mathbb{P}^{e_{j}}_{A,X}$ over $\mathcal{A} \times \mathcal{X}$. Let $\mathbb{P}^{e_{i}}_{S}$ and $\mathbb{P}^{e_{j}}_{S}$ be the induced distributions over {$\mathbb{R}^s$} by mapping function $g: \mathcal{A} \times \mathcal{X} \rightarrow \mathbb{R}^s$, then we have:
\begin{equation*}
% \footnotesize
    d_{JS}(\mathbb{P}^{e_{i}}_{A,X}, \mathbb{P}^{e_{j}}_{A,X}) \geq d_{JS}(\mathbb{P}^{e_{i}}_{S}, \mathbb{P}^{e_{j}}_{S})
\end{equation*}
\end{lemma}
\begin{lemma}\label{lemma:2}
$\forall i,j$, JS distance between $\mathbb{P}^{e_{i}}_{S,Y}$ and $\mathbb{P}^{e_{j}}_{S,Y}$ in  Eq. \eqref{eq:upper_acc}  can be decomposed:
\begin{align}
% \footnotesize
 d_{JS}\left(\mathbb{P}^{e_{i}}_{S,Y}, \mathbb{P}^{e_{j}}_{S,Y}\right) =&  d_{JS}\left(\mathbb{P}^{e_{i}}_Y, \mathbb{P}^{e_{j}}_Y\right) \nonumber\\&+ \sqrt{2\mathbb{E}_{{y\sim \mathbb{P}^{e_{i,j}}_Y}} \left[ {d}_{JS} \left(\mathbb{P}^{e_{i}}_{S|Y}, \mathbb{P}^{e_{j}}_{S|Y} \right)^2 \right] } 
\end{align}
where {$\mathbb{P}^{e_{i,j}}_Y = \frac{1}{2} \left (\mathbb{P}^{e_{i}}_Y + \mathbb{P}^{e_{j}}_Y \right )$}.
\end{lemma}
Based on Lemma~\ref{lemma:1} and Lemma \ref{lemma:2}, we can integrate Inequality \eqref{eq-s1} and Inequality \eqref{eq-s2} as follows:
\begin{align}
% \footnotesize
&\epsilon_{\texttt{Acc}}^{e_{T}}\left(\hat{f} \circ g\right) \leq \underbrace{\frac{1}{K}\sum_{i=1}^{K} \epsilon_{\texttt{Acc}}^{e_{i}}\left(\hat{f} \circ g\right)}_{\textbf{term (i)}} + \underbrace{\sqrt{2\mathbb{E}_{{y\sim \mathbb{P}^{e_{i,j}}_Y}} \left[ {d}_{JS} \left(\mathbb{P}^{e_{i}}_{S|Y}, \mathbb{P}^{e_{j}}_{S|Y} \right)^2 \right] }}_{\textbf{term (ii)}} \nonumber \\
&+ \sqrt{2}\pi_c \underset{i \in [K]}{\max} d_{JS}\left(\mathbb{P}^{e_{T}}_{A,X,Y}, \mathbb{P}^{e_{i}}_{A,X,Y}\right) + \sqrt{2} \pi_c \underset{i,j \in [K]}{\max} d_{JS}\left(\mathbb{P}^{e_{i}}_Y, \mathbb{P}^{e_{j}}_Y\right),
\end{align}
and this completes the proof.
\end{proof}

\subsection{Proof for Theorem~\ref{theorem-2}}
\begin{proof}

\begin{lemma}\label{lemma:3}
The following holds for any domain $e$:
$$\sqrt{\epsilon_{\texttt{Acc}}^{e}(\hat{f} \circ g)} = \sqrt{\mathbb{E}_{e}[\mathcal{L}(\hat{f} \circ g(A,X),Y)]}\geq \sqrt{\frac{2\pi_c}{\xi}}d_{JS}(\mathbb{P}_Y^{e},\mathbb{P}^{e}_{\hat{Y}})^2, ~\forall \hat{f} \circ g $$
where $\hat{Y}$ is the prediction made by randomized predictor $\hat{f} \circ g$. 
\end{lemma}
Consider a source domain $e_i$ and target domain $e_T$. Because JS-distance $d_{JS}(\cdot,\cdot)$ is a distance metric, we have triangle inequality:
$$d_{JS}(\mathbb{P}^{e_i}_{Y},\mathbb{P}^{e_T}_{Y})\leq d_{JS}(\mathbb{P}^{e_i}_{Y},\mathbb{P}^{e_i}_{\hat{Y}}) + d_{JS}(\mathbb{P}^{e_i}_{\hat{Y}},\mathbb{P}^{e_T}_{\hat{Y}}) + d_{JS}(\mathbb{P}^{e_T}_{\hat{Y}},\mathbb{P}^{e_T}_{Y}) $$

Since $A, X\overset{g}{\longrightarrow} S\overset{\hat{f}}{\longrightarrow} \hat{Y}$, we have $d_{JS}(\mathbb{P}^{e_i}_{\hat{Y}},\mathbb{P}^{e_T}_{\hat{Y}})\leq d_{JS}(\mathbb{P}^{e_i}_{S},\mathbb{P}^{e_T}_{S})\leq(\mathbb{P}^{e_i}_{A,X},\mathbb{P}^{e_T}_{A,X})$. Using Lemma \ref{lemma:3}, the bound holds as follows:
\begin{align}
% \footnotesize
 &\Big(d_{JS}(\mathbb{P}^{e_i}_{Y},\mathbb{P}^{e_T}_{Y}) -d_{JS}(\mathbb{P}^{e_i}_{X},\mathbb{P}^{e_T}_{X})\Big)^2\nonumber\\
 \leq&  \Big(d_{JS}(\mathbb{P}^{e_i}_{Y},\mathbb{P}^{e_i}_{\hat{Y}}) + d_{JS}(\mathbb{P}^{e_T}_{\hat{Y}},\mathbb{P}^{e_T}_{Y}) \Big)^2 \nonumber\\
 \leq & 2\Big(d_{JS}(\mathbb{P}^{e_i}_{Y},\mathbb{P}^{e_i}_{\hat{Y}})^2 +  d_{JS}(\mathbb{P}^{e_T}_{\hat{Y}},\mathbb{P}^{e_T}_{Y})^2\Big)\nonumber\\
 \leq& \frac{2}{\sqrt{\frac{2\pi_c}{\xi}}}\Big( \sqrt{\epsilon_{\texttt{Acc}}^{e_i}(\hat{f} \circ g)}+\sqrt{\epsilon_{\texttt{Acc}}^{e_T}(\hat{f} \circ g)}\Big)\nonumber\\
 \leq& \sqrt{\frac{4\xi}{\pi_c}\Big(\epsilon_{\texttt{Acc}}^{e_i}(\hat{f} \circ g)+\epsilon_{\texttt{Acc}}^{e_T}(\hat{f} \circ g)\Big)}
\end{align}

%\begin{eqnarray*}
% \sqrt{c}\Big(d_{JS}(\mathbb{P}^{e_i}_{Y},\mathbb{P}^{e_T}_{Y})^2- d_{JS}(\mathbb{P}^{e_i}^{Z},\mathbb{P}^{e_T}^{Z})^2\Big) &\leq& \sqrt{\epsilon^{\texttt{Acc}}_{e_i^S}(\hat{f} \circ g)} + \sqrt{\epsilon^{\texttt{Acc}}_{e_T}(\hat{f} \circ g)}\\
%&\leq& \sqrt{2\Big(\epsilon^{\texttt{Acc}}_{e_i^S}(\hat{f} \circ g)+\epsilon^{\texttt{Acc}}_{e_T}(\hat{f} \circ g)\Big)} 
%\end{eqnarray*}
The last inequality is AM-GM inequality. 

Therefore, we have
$$\epsilon_{\texttt{Acc}}^{e_i}(\hat{f} \circ g)+\epsilon_{\texttt{Acc}}^{e_T}(\hat{f} \circ g)\geq  \frac{\pi_c}{4\xi}\Big(d_{JS}(\mathbb{P}^{e_i}_{Y},\mathbb{P}^{e_T}_{Y})- d_{JS}(\mathbb{P}^{e_i}_{A,X},\mathbb{P}^{e_T}_{A,X}) \Big)^4$$.
The above holds for any source domain $e_i$. Average overall $K$ source domains, we have
\begin{align}
% \footnotesize
\label{eq:lower-1}
    &\frac{1}{K}\sum_{i=1}^K\epsilon_{\texttt{Acc}}^{e_i}(\hat{f} \circ g)+\epsilon_{\texttt{Acc}}^{e_T}(\hat{f} \circ g)\nonumber\\
    \geq&  \frac{\pi_c}{4\xi K}\sum_{i=1}^K\Big(d_{JS}(\mathbb{P}^{e_i}_{Y},\mathbb{P}^{e_T}_{Y})- d_{JS}(\mathbb{P}^{e_i}_{A,X},\mathbb{P}^{e_T}_{A,X}) \Big)^4.
\end{align}
\end{proof}

\section{More Experiment Results}
\subsection{Sensitivity Analysis.}
\label{Sensitivity}
\begin{figure}[t]
	\centering
	\begin{minipage}[b]{0.485\textwidth}
    \centering
		\includegraphics[width=\linewidth]{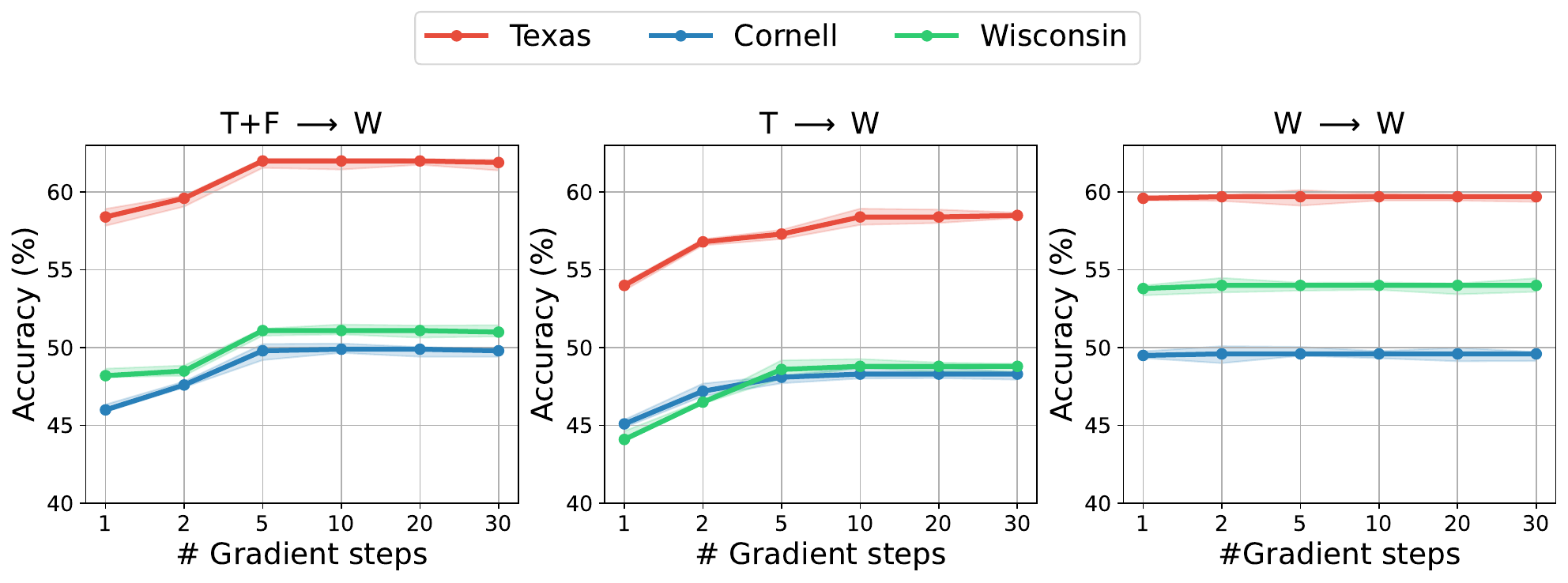}
		\caption{Accuracy with different gradient steps during testing.}
   % \vspace{-3mm}
		\label{update_test}
	\end{minipage}
	\hfill
	\begin{minipage}[b]{0.485\textwidth}
		\centering
		\includegraphics[width=\linewidth]{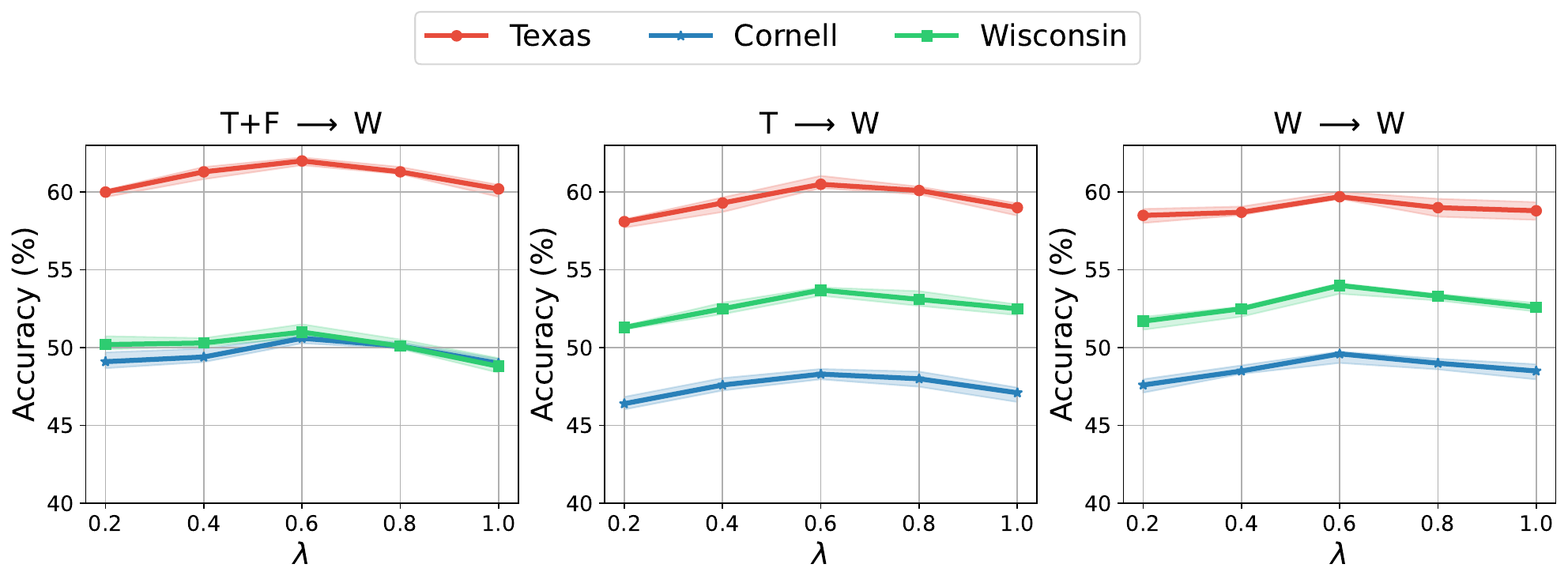}
		\caption{Accuracy with different weight of input graph $\lambda$.}
  % \vspace{-3mm}
		\label{lamda}
	\end{minipage}
\end{figure}

% We further evaluate the sensitivity of \sysname{} concerning the $h$ (the number of hub nodes), $\mathcal{L}^{rec}$ (the weight coefficient of ) and $\mathcal{L}^{elbo}$.
We evaluate the sensitivity of \sysname{} to varying numbers of gradient steps during meta-testing and the weight of the original graph $\lambda$.
The results are shown in Fig~\ref{update_test} and Fig~\ref{lamda}.
Owing to space constraints, we report the results for three graphs in \textsc{WebKB} across the three scenarios.
Similar sensitivity trends are observed across other graphs as well.
We notice that in-dataset scenarios, fewer update steps are required to attain optimal results compared to cross-dataset scenarios. 
This is attributed to the greater similarity in semantic information between the test and training domains when sourced from the same dataset.
Notably, for cross-dataset scenarios, it can quickly adapt to target graphs with very little fine-tuning, and for in-dataset scenarios, it can achieve good performance even without any fine-tuning.
This phenomenon indicates that the meta-learned structure learner and representation learner can be applied to the target graph to achieve high accuracy.
Overall, the model is robust to $\lambda$.
When we use only the input graph as GNN input ($\lambda=1$), the performance degrades, indicating the importance of the structure learner.

\end{document}